\documentclass{article}

\usepackage{amssymb,amsmath,amsthm}
\usepackage{mathtools}
\mathtoolsset{showonlyrefs}

\usepackage{mathbbol}

\usepackage[final,nonatbib]{neurips_2022}

\usepackage[utf8]{inputenc} % allow utf-8 input
\usepackage[T1]{fontenc}    % use 8-bit T1 fonts
\usepackage{url}            % simple URL typesetting
\usepackage{booktabs}       % professional-quality tables
\usepackage{amsfonts}       % blackboard math symbols
\usepackage{nicefrac}       % compact symbols for 1/2, etc.
\usepackage{microtype}      % microtypography
\usepackage{xcolor}         % colors

\usepackage{enumitem}
\usepackage{wrapfig}

\DeclareMathOperator*{\argmax}{argmax}
\DeclareMathOperator*{\argmin}{argmin}
\DeclareMathOperator*{\dom}{dom}
\DeclareMathOperator*{\conv}{conv}

\def\RR{\mathbb{R}}
\def\EE{\mathbb{E}}
\def\PP{\mathbb{P}}

\def\cC{\mathcal{C}}

\def\cL{\mathcal{L}}

\def\cV{\mathcal{V}}
\def\cX{\mathcal{X}}
\def\cY{\mathcal{Y}}

\def\Reg{\Omega}
\def\bPhi{\Psi}

\usepackage{mdframed}
\definecolor{theoremcolor}{rgb}{255, 255, 255}
\mdfsetup{
    backgroundcolor=theoremcolor,
    linewidth=0.5pt,
}

\newmdtheoremenv{definition}{Definition}
\newmdtheoremenv{proposition}{Proposition}
\newmdtheoremenv{corollary}{Corollary}
\newmdtheoremenv{theorem}{Theorem}
\newmdtheoremenv{lemma}{Lemma}

\usepackage{hyperref}
\ifdefined\nohyperref\else\ifdefined\hypersetup
  \definecolor{mydarkblue}{rgb}{0,0.50,0.30}
  \hypersetup{ %
    pdftitle={},
    pdfauthor={},
    pdfsubject={},
    pdfkeywords={},
    pdfborder=0 0 0,
    pdfpagemode=UseNone,
    colorlinks=true,
    linkcolor=mydarkblue,
    citecolor=mydarkblue,
    filecolor=mydarkblue,
    urlcolor=mydarkblue,
    pdfview=FitH}

  \ifdefined\isaccepted \else
    \hypersetup{pdfauthor={Anonymous Submission}}
  \fi
\fi\fi

\title{Learning Energy Networks\\ with Generalized Fenchel-Young Losses}

\author{%
  Mathieu Blondel,  
  Felipe Llinares-L\'{o}pez,\\
  {\bf Robert Dadashi,
  L\'{e}onard Hussenot,
  Matthieu Geist} \\
  Google Research, Brain team \\
  \texttt{\{mblondel,fllinares,dadashi,hussenot,mfgeist\}@google.com} \\
}

\begin{document}

\maketitle

\begin{abstract}
Energy-based models, a.k.a.\ energy networks, perform inference by optimizing 
an energy function, typically parametrized by a neural network. 
This allows one to capture potentially complex relationships between inputs and
outputs.
To learn the parameters of the energy function, the solution to that
optimization problem is typically fed into a loss function.
The key challenge for training energy networks lies in computing loss gradients,
as this typically requires argmin/argmax differentiation.
In this paper, building upon a generalized notion of conjugate function,
which replaces the usual bilinear pairing with a general energy function,
we propose generalized Fenchel-Young losses, a natural loss construction for
learning energy networks. Our losses enjoy many desirable properties and their
gradients can be computed efficiently without argmin/argmax differentiation.
We also prove the calibration of their excess risk in the case of linear-concave
energies. We demonstrate our losses on multilabel classification and 
imitation learning tasks.
\end{abstract}

\section{Introduction}

Training a neural network usually involves finding its parameters by minimizing
a loss function, which captures how well the network fits the data. A typical
example is the multiclass logistic loss (a.k.a. cross-entropy loss)
from logistic regression, 
which is the canonical loss associated with the softmax output layer and the
categorical distribution.
If we replace the softmax with another output layer, what loss function should
we use instead? In generalized linear models \cite{glm,mccullagh_1989}, which
include logistic regression and Poisson regression as special cases,
the negative log-likelihood gives a loss function
associated with a link function, generalizing the softmax to other members
of the exponential family \cite{exponential_families}.
More generally, \textbf{Fenchel-Young losses} \cite{fylosses_jmlr} provide a
generic way to construct a canonical convex loss function if the associated
output layer can be written in a certain argmax form. 
Besides the aforementioned generalized linear models,
models that fall in this family include sparsemax \cite{sparsemax}, the
structured perceptron \cite{structured_perceptron} and conditional random fields
\cite{Lafferty2001,sutton_introduction_2011}.  However, the theory of
Fenchel-Young losses is currently limited to argmax output layers that use a
\textbf{bilinear} pairing.

To increase expressivity, energy-based models \cite{lecun_2006}, a.k.a.\
\textbf{energy networks}, perform inference by optimizing 
an \textbf{energy function},
typically parametrized by a neural network. This leads to an inner optimization
problem, which can capture potentially complex
relationships between inputs and outputs.
A similar approach is taken in SPENs (Structured Prediction Energy Networks) 
\cite{belanger_2016,belanger_2017}, in
which a continuous relaxation of the inner optimization problem is used,
amenable to projected gradient descent or mirror descent. In input-convex
neural networks \cite{icnn_icml}, energy functions are restricted to be convex,
so as to make the inner optimization problem easy to solve optimally. 
To learn the parameters of the energy function, the solution of the inner 
optimization problem is typically fed into a loss function, leading to an outer
optimization problem. In order to solve that problem by stochastic gradient
descent, the main challenge lies in computing the loss gradients. Indeed, when
using an arbitrary loss function, by the chain rule, computing gradients
requires differentiating through the inner optimization problem solution,
often referred to as \textbf{argmin} or \textbf{argmax differentiation}.
This can be done by backpropagation through unrolled
algorithm iterates~\cite{belanger_2017} or
implicit differentiation through the optimality conditions \cite{icnn_icml}.
This issue can be circumvented by using a generalized perceptron loss 
\cite{lecun_2006} or a max margin loss \cite{belanger_2016}.
These losses only require \textbf{max differentiation}, for which
\textbf{envelope theorems} \cite{milgrom_2002} can be used.  However, these
losses often fail to satisfy envelope theorem assumptions such as unicity of the
solution and lack theoretical guarantees.

\looseness=-1
In this paper, we propose to extend the theory of Fenchel-Young losses
\cite{fylosses_jmlr}, so as to create a canonical loss function associated with
an energy network. Our proposal builds upon 
\textbf{generalized conjugate functions}, 
also known as \textbf{Fenchel-Moreau conjugates}
\cite{moreau_1966,rockafellar_2009}, allowing us to go beyond bilinear pairings
and to support \textbf{general energy functions} such as neural networks. 
We introduce \textbf{regularization}, allowing us to obtain unicity of the
solutions and to use envelope theorems to compute gradients, without argmin or
argmax differentiation.
We provide novel guarantees on the \textbf{calibration} of the excess risk,
when our loss is used as a surrogate for another discrete loss.
To sum up, we obtain a well-motivated loss construction for general energy
networks. The rest of the paper is organized as follows.

\begin{itemize}[topsep=0pt,itemsep=2pt,parsep=2pt,leftmargin=10pt]

\item After providing some background (\S\ref{sec:background}), we describe
    \textbf{regularized energy networks}, identify a classification of energy
    functions and give several possible examples that fall in this family
    (\S\ref{sec:energy_networks}).

\item We define \textbf{generalized conjugates}, 
an extension of Legendre-Fenchel conjugates, and state
    their properties (\S\ref{sec:phi_convex_conjugates}).
    We establish novel conditions for a generalized conjugate to be a smooth
    function.

\item We then introduce \textbf{generalized Fenchel-Young losses} and
    show that they enjoy many of the same favorable properties as the regular
    Fenchel-Young losses (\S\ref{sec:phi_fy_losses}). 
    On the theoretical side, we prove \textbf{calibration} of the excess risk 
    for linear-concave energies and strongly-convex regularizers
    (\S\ref{sec:calibration}).

\item We demonstrate our losses on \textbf{multilabel classification} and 
    \textbf{imitation learning} tasks (\S\ref{sec:experiments}).

\end{itemize}

\section{Background}
\label{sec:background}

\paragraph{Convex conjugates.}

Let $\cC \subseteq \RR^k$ be an output set. 
Let $\Reg \colon \cC \to \RR$ be a function, such that $\Reg(p) = \infty$ 
for all $p \not \in \cC$, i.e., $\dom(\Reg) = \cC$.
Given $v \in \cV \subseteq \RR^k$, we define the convex conjugate 
\cite{boyd_book} of $\Reg$, 
also known as the Legendre-Fenchel transform of $\Reg$, by
\begin{equation}
\Reg^*(v) \coloneqq \max_{p \in \cC} ~ \langle v, p \rangle - \Reg(p).
\label{eq:convex_conjugate}
\end{equation}
The conjugate $\Reg^*$ is always convex, even if $\Reg$ is not.
We define the corresponding argmax as
\begin{equation}
p_\Reg(v) 
\coloneqq \argmax_{p \in \cC} ~ \langle v, p \rangle - \Reg(p) 
= \nabla \Reg^*(v).
\label{eq:argmax_bilinear}
\end{equation}
The latter equality follows from Danskin's theorem
\cite{danskin_1967,bertsekas_1997}, under the assumption that $\Reg$ is
strictly convex (otherwise, we obtain a subgradient).
It is well-known that $\Reg^*$ is $\frac{1}{\gamma}$-smooth w.r.t.\
the dual norm $\|\cdot\|_*$ if and only if $\Reg$ is $\gamma$-strongly
convex w.r.t.\ the norm $\|\cdot\|$ 
\cite{hiriart_1993,kakade_2009,beck_2017,zhou_2018}.

\paragraph{Fenchel-Young losses.}

Suppose that $v = g_\theta(x) \in \RR^k$ 
are the logits / scores produced by a neural network $g$,
where $x \in \cX \subseteq \RR^d$ and $\theta \in \Theta$ are the network's
input features and parameters, respectively.
What loss function should we use if we want to use 
\eqref{eq:argmax_bilinear} as output layer?
The Fenchel-Young loss \cite{fylosses_jmlr} generated 
by $\Reg \colon \cC \to \RR$ provides a natural solution. It is defined by
\begin{equation}
L_\Reg(v, y) \coloneqq \Reg^*(v) + \Reg(y) - \langle v, y \rangle,
\label{eq:fy_loss}
\end{equation}
where $y \in \cY \subseteq \cC$ is the ground-truth label.
Earlier instances of this loss were independently proposed in the 
contexts of ranking \cite{acharyya_2013} and multiclass classification
\cite{duchi_2016}.
Among many useful properties, this loss satisfies 
$L_\Reg(v, y) \ge 0$ 
and 
$L_\Reg(v, y) = 0 \Leftrightarrow y = p_\Reg(v)$ 
if $\Reg$ is strictly convex \cite{fylosses_jmlr}. 
In that sense, it is the
canonical loss associated with \eqref{eq:argmax_bilinear}.
Interestingly, many existing loss functions are recovered as special cases.
For instance, if $\cC = \RR^k$ and $\Reg(p) = \frac{1}{2} \|p\|^2_2$, 
which is $1$-strongly convex w.r.t. $\|\cdot\|_2$ over $\RR^k$,
then we obtain the self-dual, the identity mapping and the squared loss:
\begin{equation}
\Reg^*(v) = \frac{1}{2} \|v\|_2^2,
\quad
p_\Reg(v) = v,
\quad
L_\Reg(v, y) = \frac{1}{2} \|v - y\|^2_2.
\end{equation}
If $\cC$ is the probability simplex 
$\triangle^k \coloneqq \{p \in \RR^k_+ \colon \sum_{i=1}^k p_i = 1\}$ 
and 
$\Reg(p)$ is the scaled Shannon negentropy 
$\gamma \langle p, \log p \rangle$, 
which is $\gamma$-strongly convex w.r.t. $\|\cdot\|_1$ over $\triangle^k$,
then we obtain
\begin{equation}
\Reg^*(v) = \text{LSE}^\gamma(v),
\quad
p_\Reg(v) = \text{softmax}^\gamma(v),
\quad
L_\Reg(v, y) = \text{KL}(y, \text{softmax}^\gamma(v)),
\end{equation}
where we used the log-sum-exp
$\text{LSE}^\gamma(v) \coloneqq \gamma \log(\sum_{i=1}^k \exp(v_i / \gamma))$,
$\text{softmax}^\gamma(v) \propto \exp(v / \gamma)$,
and the Kullback-Leibler divergence.
More generally, when $\cC = \conv(\cY)$, the convex hull of $\cY$,
$p_\Reg$ corresponds to a projection, for which efficient
algorithms exist in numerous cases \cite{fylosses_jmlr,projection_oracles}.
The calibration of the excess risk of Fenchel-Young losses when
$\Reg$ is strongly convex was established in
\cite{nowak_2019,projection_oracles}.
However, the theory of Fenchel-Young losses is currently limited to the 
bilinear pairing $\langle v, p \rangle$, which restricts their expressivity 
and scope. 

\vspace{-0.3cm}
\section{Regularized energy networks}
\label{sec:energy_networks}

\vspace{-0.2cm}
\paragraph{Energy networks.}

\looseness=-1
Also known as energy-based models or EBMs \cite{lecun_2006},
of which SPENs \cite{belanger_2016, belanger_2017} are a particular case,
these networks compute predictions by solving an optimization problem
of the form
\begin{equation}
p^\Phi(v) \coloneqq \argmax_{p \in \cC} ~ \Phi(v, p),
\end{equation}
where $v = g_\theta(x) \in \cV$ is the energy network's input,
$\Phi(v, p)$ is a scalar-valued energy function,
and $\cC$ is an output set.
Throughout this paper, we use the convention that higher energy indicates 
higher degree of compatibility between $v$ and the prediction $p$.
Since $\Phi(v, p)$ is a general energy function, we emphasize that $v$ and $p$
do not need to have the same dimensions, unlike with the bilinear pairing
$\langle v, p \rangle$.
Any neural network $g_\theta(x) \in \RR^k$ can be written
in energy network form, since 
$\argmax_{p \in \RR^k} -\|g_\theta(x) - p\|_2^2 = g_\theta(x)$.
The key advantage of energy networks, however, is their ability to capture 
complex interactions between inputs and outputs. 
This is especially useful in the
structured prediction setting \cite{bakir_2007}, where predictions are made of
sub-parts, such as sequences being composed of individual elements.

\vspace{-0.2cm}
\paragraph{Introducing regularization.}

In this paper, we compute predictions by solving
\begin{equation}
p^\Phi_\Reg(v) 
\coloneqq \argmax_{p \in \cC} ~ \Phi(v, p) -  \Reg(p),
\label{eq:phi_convex_argmax_x}
\end{equation}
where we further added a regularization function $\Reg \colon \cC \to \RR$.
We call $\Phi(v, p) -  \Reg(p)$ a \textbf{regularized energy function}
and we call the corresponding model a \textbf{regularized energy network}.
We obviously recover usual energy networks by setting $\Reg$ to the indicator
function of the set $\cC$,
i.e., $\Omega(p) = 0$ if $p \in \cC$, $\infty$ otherwise.
While it is in principle possible to absorb $\Reg$ into $\Phi$,
keeping $\Reg$ explicit has several advantages: 
1) it allows to introduce generalized conjugate
functions and their properties (\S\ref{sec:phi_convex_conjugates}) 
2) it allows to introduce generalized Fenchel-Young losses
(\S\ref{sec:phi_fy_losses}), which mirror the original
Fenchel-Young losses 
3) we obtain closed forms for \eqref{eq:phi_convex_argmax_x} in certain cases 
(Table \ref{tab:network_examples}).

\paragraph{Solving the maximization problem.}

The ability to solve the maximization problem in \eqref{eq:phi_convex_argmax_x}
efficiently depends on the properties of $\Phi(v, p)$ w.r.t.\ $p$. 
If $\Phi(v, p)$ is \textbf{linear} in $p$, for instance
$\Phi(v, p) = \langle v, U p \rangle$,
then we obtain
$p^\Phi_\Reg(v) = p_\Reg(U^\top v)$.
Thus, the computation of \eqref{eq:phi_convex_argmax_x} reduces to
\eqref{eq:argmax_bilinear}, for which closed forms are often available.
If $\Phi(v, p)$ is \textbf{concave} in $p$ and $\cC$ is a convex set, 
then we can solve
\eqref{eq:phi_convex_argmax_x} in polynomial time 
using an iterative algorithm, such as projected gradient ascent.
This is the most general energy class for which
our loss and its gradient can be computed to arbitrary precision.
If $\Phi(v, p)$ is nonconcave in $p$, then it is not possible to solve
\eqref{eq:phi_convex_argmax_x} in polynomial time in general, unless $\cC$ is a
discrete set of small cardinality.
In general, we emphasize that $\Phi(v, p)$ can be nonconvex in $v$,
as is typical with neural networks, since the maximization problem in
\eqref{eq:phi_convex_argmax_x} is w.r.t. $p \in \cC$.
However, certain properties we establish later
require $\Phi(v, p)$ to be convex in $v$.
We now give examples of regularized energy networks in increasing order of
expressivity / complexity.

\paragraph{Generalized linear models.} 

As a warm up, we consider the case of \textbf{bilinear} energy 
$\Phi(v, p)$.
For instance, for probabilistic classification with $k$
classes, where the goal is to predict $p \in \triangle^k$ from $x \in \RR^d$,
we can use
$\Phi(v, p) = \langle v, p \rangle$ with $v = W x + b$,
where $W \in \RR^{k \times d}$ and $b \in \RR^k$.
In this case, we therefore obtain
$p^\Phi_\Reg(v) = p_\Reg(Wx+b)$,
recovering generalized linear models \cite{glm,mccullagh_1989}, the
structured perceptron \cite{structured_perceptron} and conditional random fields
\cite{Lafferty2001,sutton_introduction_2011}.

\paragraph{Rectifier and maxout networks.}

We now consider the case of \textbf{convex-linear} energy $\Phi(v, p)$.
A first example is a rectifier network \cite{glorot_2011} with one hidden layer.
Indeed, $\Phi(v, p) = \langle \sigma(v), U p \rangle$ is convex in $v$
if $\sigma$ is an element-wise, convex and non-decreasing activation function, 
such as the relu or softplus, 
and if $U$ and $p$ are non-negative. 
A second example is a maxout network \cite{maxout} (i.e., a max-affine
function) or its smoothed counterpart, the log-sum-exp network
\cite{calafiore_2019}. Indeed, $\Phi(v, p)
= \sigma(v) \cdot p$, where $\sigma$ is the max or log-sum-exp operator,
is convex in $v$, if the scalar $p$ is non-negative. 
In general, it is always possible to construct a \textbf{convex-concave} energy
from a jointly-convex function
using the Legendre-Fenchel transform in the first argument.

\paragraph{Input-convex neural networks.} 

As an example of \textbf{nonconvex-concave} energy,
we consider
$\Phi(v, p) = -\text{ICNN}(v, p)$, 
where $\text{ICNN}$ is an input-convex neural network \cite{icnn_icml},
i.e., it is convex in $p$ but can be nonconvex in $v$.
If $\Reg(p)$ and $\cC$ are convex, then $\Phi(v, p)$ is concave in $p$ and 
\eqref{eq:phi_convex_argmax_x} can be solved in polynomial time
using an iterative algorithm, such as projected gradient ascent.

\paragraph{Probabilistic energy networks.}

It is often desirable to define a conditional probability distribution
\begin{equation}
\PP(Y=y|X=x) \coloneqq 
\frac{\exp(E(v, y))}{\sum_{y' \in \cY} \exp(E(v, y'))}.
\end{equation}
Such networks are typically trained using a cross-entropy loss (or equivalently,
negative log-likelihood).
Unfortunately, when $\cY$ is large or infinite, this loss and its gradients are
intractable to compute, due to the normalization constant.
Therefore, recent research has been devoted to developing approximate
training schemes \cite{song_2021, no_mcmc, nash_2019}. Although this is not the
focus of this paper, we point out that probabilistic energy networks can also be
seen as regularized energy networks in the space of probability distributions
$\cC = \triangle^{|\cY|}$, 
if we set $\Phi(v, p) = \sum_{y \in \cY} p(y) E(v, y)$
and $\Reg(p) = \sum_{y \in \cY} p(y) \log p(y)$.
%\begin{equation}
%\small
%p_\Reg^\Phi(v)
%=
%\PP(Y=\cdot|X=x) 
%= \argmax_{p \in \triangle^{|\cY|}}
%\Phi(v, p) - \Reg(p)
%\coloneqq \argmax_{p \in \triangle^{|\cY|}}
%\sum_{y \in \cY} p(y) E(v, y) 
%- \sum_{y \in \cY} p(y) \log p(y).
%\end{equation}
While the resulting optimization problem is intractable in general,
it does suggest possible approximation schemes, such as the use of Frank-Wolfe
methods \cite{belanger_2013,barrierfw,sparsemap}.

\paragraph{Existing loss functions for energy networks.} 
Given a pair $(x, y)$ and output $v = g_\theta(x)$,
how do we measure the discrepancy between $p^\Phi_\Reg(v)$ and $y$?
One possibility \cite{icnn_icml,belanger_2017}
is to use the composition of a differentiable loss 
$L \colon \cC \times \cY \to \RR_+$
with the argmax output, namely $(v, y) \mapsto L(p^\Phi_\Reg(v), y)$.
However, computing the loss gradient w.r.t.\ $v$ then requires 
to differentiate through $p^\Phi_\Reg(v)$,
either through unrolling or implicit differentiation.
This is particularly problematic when
$\cC$ is a complicated convex set, as differentiating through a
projection can be challenging. In contrast, our proposed loss completely
circumvents this need and enjoys easy-to-compute gradients.
For unregularized energy networks,
a naive idea, called the energy loss, is to use $(v, y)
\mapsto -\Phi(v, y)$. However, this loss only works well if $\Phi$ is a
similarity measure and works poorly in general \cite{lecun_2006}.
A better choice
is the generalized perceptron loss $(v, y) \mapsto \max_{p \in \cC} \Phi(v, p) -
\Phi(v, y)$ \cite{lecun_2006}.
Our loss can be seen as a principled generalization of this loss to
regularized energy networks, with theoretical guarantees.

\begin{table}[t]
\caption{Examples of regularized energy networks. 
    The $v$ and $p$ columns indicate the
    property of the energy function $\Phi(v, p)$ in these variables. The
$L_\Reg^\Phi(v, y)$ column indicates the property of the loss in $v$. 
The linear-quadratic energy uses $v = (A, b)$ where $A$ is negative
semi-definite and $\Reg(p) = \frac{\gamma}{2} \|p\|^2_2$.
%The abbreviation ``DC'' means ``difference of convex functions''. 
%We recall that a linear function is both convex and concave.
}
\begin{center}
\begin{small}
\begin{tabular}{cccccc}
\toprule
& $\Phi(v, p)$ & $v$ & $p$ & $L^\Phi_\Reg(v, y)$ & $p_\Reg^\Phi(v)$ \\
\midrule
GLM & $\langle v, p \rangle$ & linear & linear & convex & 
$p_\Reg(v)$ \\
Linear-quadratic & 
$\frac{1}{2} \langle p, Ap\rangle + \langle p, b \rangle$ & 
linear & quadratic & convex &
$(\gamma I - A)^{-1} b$ \\
Rectifier network & $\langle \text{relu}(v), Up \rangle$ & convex & linear & 
DC & $p_\Reg(U^\top \text{relu}(v))$ \\
Maxout network & $p \cdot \text{max}(v)$ & convex & linear & 
DC & $p_\Reg(\text{max}(v))$ \\
LSE network & $p \cdot \text{LSE}^\gamma(v)$ & convex & linear & 
DC & $p_\Reg(\text{LSE}^\gamma(v))$ \\
ICNN & $-\text{ICNN}(v, p)$ & nonconvex & concave & nonconvex & 
no closed form \\
Probabilistic & $\sum_{y \in \cY} p(y) E(v, y)$ & nonconvex & linear & 
nonconvex & 
$\frac{\exp(E(v, \cdot))}{\sum_{y' \in \cY} \exp(E(v, y'))}$ \\
Arbitrary & $\Phi(v, p)$ & nonconvex & nonconcave & nonconvex & 
no closed form \\
\bottomrule
\end{tabular}
%{\tiny GLM: generalized linear model. DC: difference of convex functions. ICNN:
%input-convex neural network.  LSE: log-sum-exp.}
\end{small}
\end{center}
\label{tab:network_examples}
\end{table}

\vspace{-0.4cm}
\section{Generalized conjugates}
\label{sec:phi_convex_conjugates}

In order to devise generalized Fenchel-Young losses, 
we build upon a generalization due to Moreau \cite[Chapter
14]{moreau_1966} of the convex conjugate $\Reg^*$. Denoted $\Reg^\Phi$, 
it replaces
the bilinear pairing in \eqref{eq:convex_conjugate} with a more general coupling
$\Phi$ \cite[Chapter 11, Section L]{rockafellar_2009}. In this section, we state
their definition, properties, closed-form expressions and connection to the
$C$-transform in optimal transport.

\paragraph{Definition.}

Let $\Phi(v, p) \in \RR$ be a coupling / energy function. 
The $\Phi$\textbf{-convex conjugate} of $\Reg \colon \cC
\to \RR$, also known as \textbf{Fenchel-Moreau conjugate}, is then defined by
the value function
\begin{equation}
\Reg^\Phi(v) \coloneqq \max_{p \in \cC} ~ \Phi(v, p) -  \Reg(p).
\label{eq:phi_convex_conjugate}
\end{equation}
The $\Phi$-convex conjugate is an important tool in abstract
convex analysis \cite{singer_1997,rubinov_2000}.
Recently, it has been used to provide 
``Bellman-like'' equations in stochastic
dynamic programming \cite{chancelier_2018} and to provide tropical 
analogues of reproducing kernels \cite{aubin_2022}. 
We assume that the maximum is feasible for all $v \in \cV = \RR^d$, 
meaning that $\dom(\Reg^\Phi) = \cV$. We emphasize again that,
unlike with \eqref{eq:convex_conjugate},
$v$ and $p$ do not need to have compatible dimensions.
We denote the argmax solution corresponding to \eqref{eq:phi_convex_conjugate} 
by 
\begin{equation}
p^\Phi_\Reg(v) 
\coloneqq \argmax_{p \in \cC} ~ \Phi(v, p) - \Reg(p).
\label{eq:phi_convex_argmax}
\end{equation}
If a function $F(v)$ can be written as $F(v) = \Reg^\Phi(v)$
for some $\Reg$, it is called $\Phi$\textbf{-convex} (in analogy, a function
$f(v)$ is convex and closed if and only if it can be written as $f(v) =
\Reg^*(v)$ for some $\Reg$).

\paragraph{Properties.}

$\Phi$-convex conjugates enjoy many useful properties, some of them are natural
extensions of the usual convex conjugate properties.
Proofs are provided in Appendix \ref{proof:properties_phi_convex_conjugate}.
\begin{proposition}[Properties of $\Phi$-convex conjugates]

Let $\Reg \colon \cC \to \RR$ and $\Phi \colon \cV \times \cC \to \RR$.
\begin{enumerate}[topsep=0pt,itemsep=2pt,parsep=2pt,leftmargin=10pt]

\item \normalfont{\textbf{Generalized Fenchel-Young inequality:}}
for all $v \in \cV$ and $p \in \cC$,
\begin{equation}
\Reg^\Phi(v) + \Reg(p) - \Phi(v, p) \ge 0.
\end{equation}

\item \normalfont{\textbf{Convexity:}} 
If $\Phi(v, p)$ is convex in $v$, then $\Reg^\Phi(v)$ is convex
(even if $\Reg(p)$ is nonconvex).

\item \normalfont{\textbf{Order reversing:}} 
if $\Reg(p) \le \Lambda(p)$ for all $p \in \cC$, 
then $\Reg^\Phi(v) \ge \Lambda^\Phi(v)$ for all $v \in \cV$.

\item \normalfont{\textbf{Continuity:}} 
$\Reg^\Phi$ shares the same continuity modulus as $\Phi$.

\item \label{item:phi_conjugate_gradient} \textbf{Gradient (envelope theorem):}
Under assumptions (see paragraph below), 
we have\\ 
$\nabla \Reg^\Phi(v) = \nabla_1 \Phi(v, p^\Phi_\Reg(v))$, 
where $\nabla_1$ denotes the gradient in the first argument.

\item \label{item:smoothness} \normalfont{\textbf{Smoothness:}}
If $\cC$ is a compact convex set, $\Phi(v, p)$ is $\beta$-smooth in $(v, p)$, 
concave in $p$ and
$\Omega(p)$ is $\gamma$-strongly convex in $p$, 
then $\Reg^\Phi(v)$ is $(\beta+\beta^2/\gamma)$-smooth
and $p^\Phi_\Reg(v)$ is $\beta/\gamma$-Lipschitz.

\end{enumerate}
\label{prop:properties_phi_convex_conjugate}
\end{proposition}
The condition on $\Phi$ and $\Omega$ in item \ref{item:smoothness} 
for $\Reg^\Phi$ to be a smooth function (i.e., with Lipschitz-continuous
gradients)
is a novel result and will play a crucial role for
establishing calibration guarantees in \S\ref{sec:calibration}.

%\paragraph{Envelope theorems.}
\paragraph{Assumptions for envelope theorems.}

The expression in item \ref{item:phi_conjugate_gradient}
allows to compute $\nabla \Reg^\Phi(v)$ 
without \textit{argmax} differentiation. 
It is based on envelope theorems, 
which can be used for \textit{max} differentiation.
Indeed, we have $\Reg^\Phi(v) = \max_{p \in \cC} F(v, p)$, 
where $F(v, p) \coloneqq \Phi(v, p) - \Reg(p)$. 
We assume that the maximum is unique and $\cC$ is a compact.
If $F(v, p)$ is convex in $v$,
%(but not necessarily differentiable in $v$), 
we apply Danskin's theorem \cite{danskin_1967}
\cite[Proposition B.25]{bertsekas_1997}.
Without convexity assumption in $v$, 
if $F(v, p)$ is
continuously differentiable in $v$ for all $p \in \cC$,
$\nabla_1 F$ is continuous, we apply \cite[Theorem 10.31]{rockafellar_2009}. 
%See also \cite[Theorem 10.58]{rockafellar_2009} for a minimum counterpart.
%In practice, this is implemented using the \texttt{stop\_gradient} 
%function available in JAX \cite{jax}.
When we do not compute the exact solution of
\eqref{eq:phi_convex_argmax}, we only obtain an approximation of the gradient
$\nabla \Reg^\Phi(v)$; see \cite{ablin_2020} for approximation guarantees.
For other envelope theorem usecases in machine learning, see,
e.g., \cite{bottou_2018}.

\paragraph{Closed forms.}

While \eqref{eq:phi_convex_conjugate} and
\eqref{eq:phi_convex_argmax} may need to be solved numerically in general,
they enjoy closed-form expressions in simple cases.
Proofs are provided in Appendix \ref{proof:closed_forms}.
\begin{proposition}[Closed-form expressions]

Let $\Reg \colon \cC \to \RR$ and $\Phi \colon \cV \times \cC \to \RR$.
\begin{enumerate}[topsep=0pt,itemsep=2pt,parsep=2pt,leftmargin=10pt]

\item \normalfont{\textbf{Bilinear coupling:}}
If $\Phi(v, p) = \langle v, U p \rangle$, 
then $\Reg^\Phi(v) = \Reg^*(U^\top v)$
and $p_\Reg^\Phi(v) = p_\Reg(U^\top v)$.

\item \normalfont{\textbf{Linear-quadratic coupling:}}
If $\cC = \RR^k$, $\Omega(p) = \frac{\gamma}{2} \|p\|_2^2$ and 
$\Phi(v, p) = \frac{1}{2} \langle p, Ap\rangle + \langle p, b \rangle$,
where $v = (A, b)$ and $A$ is such that $(\gamma I - A)$ is positive definite, 
we obtain
\begin{equation}
\Reg^\Phi(v) = \frac{1}{2} \langle b, (\gamma I - A)^{-1} b \rangle
\quad \text{and} \quad
p^\Phi_\Reg(v) = (\gamma I - A)^{-1} b.
\label{eq:closed_form_qp}
\end{equation}

\item \normalfont{\textbf{Metric coupling:}}
If $\cV = \cC$, $\Phi = -C$ where $C(v, p)$ is a metric
and $\Reg$ is $M$-Lipschitz with $M \le 1$, then
$\Reg^\Phi = -\Reg$ and $p^\Phi_\Reg(v) = v$.

\end{enumerate}
\label{prop:closed_forms}
\end{proposition}

\paragraph{Relation with the $C$-transform.}

Given a cost function $C$,
we may define a $\min$ counterpart of \eqref{eq:phi_convex_conjugate},
\begin{equation}
\Lambda_C(v) \coloneqq \min_{p \in \cC} ~ C(v, p) - \Lambda(p).
\label{eq:c_transform}
\end{equation}
In the optimal transport literature, this is known as the $C$-transform of
$\Lambda$ \cite{santambrogio_2015,peyre_2019}.
When $C$ is bilinear, this recovers the notion of concave conjugate
\cite{borwein_2006}.
When $C(v, p) = c(v - p)$ for some $c$, this recovers the 
infimal convolution \cite{borwein_2006}.
It is easy to check that $\Reg = -\Lambda \Leftrightarrow \Reg^\Phi =
-\Lambda_C$ with $C = -\Phi$. Thus, $\Reg^\Phi$ and $\Lambda_C$ are the natural
extensions of convex and concave conjugates, respectively.  We opt for the
former in this paper to closely mirror the usual convex conjugates and
Fenchel-Young losses.

\vspace{-.3cm}
\section{Generalized Fenchel-Young losses}
\label{sec:phi_fy_losses}

\paragraph{Definition.}

The generalized Fenchel-Young inequality in Proposition
\ref{prop:properties_phi_convex_conjugate} leads us to propose the
generalized Fenchel-Young loss,
the natural extension of \eqref{eq:fy_loss} to $\Phi$-convex conjugates:
\begin{equation}
L_\Reg^\Phi(v, y) 
\coloneqq \Reg^\Phi(v) + \Reg(y) - \Phi(v, y).
\label{eq:phi_fy_loss}
\end{equation}
We also have the relationships
$L_\Reg^\Phi(v, y) 
= F(v, y) - F(v, p^\Phi_\Reg(v))$,
where $F(v, p) = \Reg(p) - \Phi(v, p)$,
and
$p^\Phi_\Reg(v) = \argmin_{p \in \cC} ~ L_\Reg^\Phi(v, p)$.
We now give an intuitive geometric interpretation of \eqref{eq:phi_fy_loss}.

\paragraph{Geometric interpretation.}

\begin{wrapfigure}[13]{r}{0.45\textwidth}
\centering
\vspace{-0.3cm}
\includegraphics[scale=0.95]{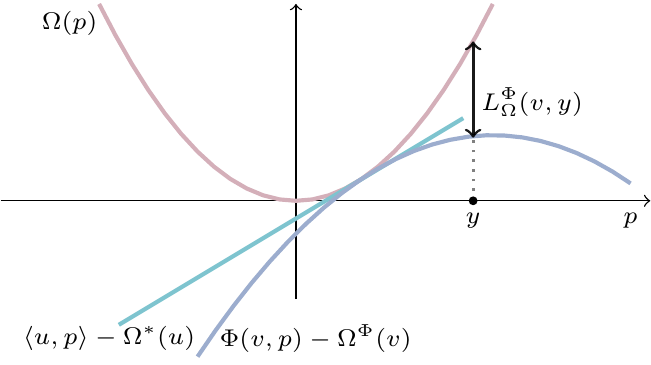}\\
\caption{Geometric interpretation with
$\Reg(p) = \frac{\gamma}{2} p^2$
and
$\Phi(v, p) = \frac{1}{2} ap^2 + bp$.
}
\label{fig:illustration}
\end{wrapfigure}
For the sake of illustration, let us set
$\Reg(p) = \frac{\gamma}{2} p^2$
and
$\Phi(v, p) = \frac{1}{2} ap^2 + bp$.
From \eqref{eq:convex_conjugate},
the line $p \mapsto \langle u, p \rangle - \Reg^*(u)$, where $\Reg^*(u) =
\frac{1}{2\gamma} u^2$, is the tightest linear lower
bound on $\Reg(p)$, here depicted with $u=0.6$.
It is the tangent of $\Reg(p)$ at $p=p_\Reg(u)$. 
Similarly, from \eqref{eq:phi_convex_conjugate}, $p \mapsto
\Phi(v, p) - \Reg^\Phi(v)$ is the tightest lower-bound of this form given $v$. 
It touches $\Reg(p)$ at $p = p^\Phi_\Reg(v)$, here depicted with $v = (a,
b)$, $a=-1$ and $b=1$, for which we have
the closed form $\Reg^\Phi(v) = \frac{1}{2}(\gamma - a)^{-1} b^2$ (Proposition
\ref{prop:closed_forms}). 
The generalized Fenchel-Young loss \eqref{eq:phi_fy_loss} is then the
\textbf{gap} between $\Omega(p)$ and $\Phi(v, p) - \Reg^\Phi(v)$, 
evaluated at the ground-truth $p=y$.  
The goal of training is to adjust the parameters $\theta$
of a network $v = g_\theta(x)$ so as to minimize this gap averaged over all
$(x,y)$ training pairs.

\paragraph{Properties.}

Generalized Fenchel-Young losses enjoy many desirable properties,
as we now show.
\begin{proposition}[Properties of generalized F-Y losses]
Let $\Reg \colon \cC \to \RR$ and $\Phi \colon \cV \times \cC \to \RR$.
\begin{enumerate}[topsep=0pt,itemsep=2pt,parsep=2pt,leftmargin=10pt]

\item \normalfont{\textbf{Non-negativity:}}
$L_\Reg^\Phi(v, p) \ge 0$ for all $v \in \cV$ and $p \in \cC$.

\item \normalfont{\textbf{Zero loss:}}
if the maximum in \eqref{eq:phi_convex_conjugate} exists and is unique,
$L_\Reg^\Phi(v, y) = 0 \Leftrightarrow y = p^\Phi_\Reg(v)$.

\item \normalfont{\textbf{Gradient:}}
$\nabla_1 L^\Phi_\Reg(v, p) 
= \nabla \Reg^\Phi(v) - \nabla_1 \Phi(v, p)$.

\item \normalfont{\textbf{Difference of convex (DC):}}
if $\Phi(v, p)$ is convex in $v$,
then $L^\Phi_\Reg(v, p)$ is a difference of convex functions in $v$:
$\Reg^\Phi(v)$ and $\Phi(v, p)$.
If $\Phi(v, p)$ is linear in $v$, then $L^\Phi_\Reg(v, p)$ is
convex in $v$.

%\item \normalfont{\textbf{Linear coupling in $p$:}}
%If $\Phi(v, p) = \langle v, U p \rangle$, then
%$L^\Phi_\Reg(v, p) = L_\Reg(U^\top v, p)$.

\item \label{item:smaller_better} 
\normalfont{\textbf{Smaller output set, smaller loss.}}
If $\cC' \subseteq \cC$ and $\Reg'$ is the restriction of $\Reg$ to $\cC'$, 
then $L^\Phi_{\Reg'}(v, p) \le L^\Phi_\Reg(v, p)$ 
for all $v \in \cV$ and $p \in \cC'$.

\item \normalfont{\textbf{Quadratic lower-bound.}}
If $\Phi(v, p) - \Reg(p)$ is $\gamma$-strongly concave in $p$ 
w.r.t. $\|\cdot\|$ over $\cC$
and $\cC$ is a closed convex set, then for all $v \in \cV$
and $p \in \cC$
\begin{equation}
\frac{\gamma}{2} \|p - p^\Phi_\Reg(v)\|^2 \le L_\Reg^\Phi(v, p).
\label{eq:quadratic_lower_bound}
\end{equation}

\item \normalfont{\textbf{Upper-bounds.}}
If $\Phi(v, p) - \Omega(p)$ is $\alpha$-Lipschitz
in $p$ w.r.t. $\|\cdot\|$ over $\cC$, then
$L^\Phi_\Reg(v, p) \le \alpha \|p - p^\Phi_\Reg(v)\|$.
If $\Phi(v, p)$ is concave in $p$, then
$L^\Phi_\Reg(v, p) \le L_\Reg(\nabla_2 \Phi(v, p), p)$.

\end{enumerate}
\label{prop:properties_phi_fy_losses}
\end{proposition}
Proofs are given in Appendix \ref{proof:properties_phi_fy_losses}.
The \textbf{non-negativity} is a good property for a loss function.
The \textbf{zero-loss} property $L_\Reg^\Phi(v, y) = 0 \Leftrightarrow 
y = p^\Phi_\Reg(v)$, which is true for instance if $p \mapsto \Phi(v, p) -
\Reg(p)$ is strictly concave, is key as it allows to use 
$p^\Phi_\Reg(v)$ defined in
\eqref{eq:phi_convex_argmax} as the (implicit)
output layer associated with an energy network.
The \textbf{gradient} $\nabla_1 L^\Phi_\Reg(v, y)$ does
not require to differentiate through the argmax problem in
\eqref{eq:phi_convex_argmax} needed for computing $p^\Phi_\Reg(v)$.
Typically, differentiating through an argmax or argmin, as is done in
input-convex neural networks \cite{icnn_icml}, requires either unrolling or
implicit differentiation
\cite{griewank_2008,bell_2008,krantz_2012,bonnans_2013,blondel_implicit_diff}
and is therefore more costly. 

The fact that $L^\Phi_\Reg(v, p)$ is a \textbf{difference of convex (DC)}
functions in $v$ when $\Phi(v, p)$ is convex in $v$ suggests that we can use DC
programming techniques, such as the convex-concave procedure \cite{yuille_2003},
for training such energy networks. We leave the investigation of this
observation to future work. The \textbf{``smaller output set, smaller loss''}
property means that we can achieve the smallest loss by choosing the smallest
set $\cC$ in \eqref{eq:phi_convex_argmax} such that $\cY \subseteq \cC$. 
The smallest such convex set is the convex hull of $\cY$, also known as marginal
polytope \cite{wainwright_2008} when $\cY \subseteq \{0,1\}^k$.
The \textbf{quadratic lower-bound} relates our loss to
using $p^\Phi_\Reg(v)$ within a squared norm loss.
The \textbf{upper-bounds} relate our loss to
using $p^\Phi_\Reg(v)$ within a norm loss and to using a regular Fenchel-Young
loss with a linearized energy.

If $\Reg$ is the indicator function of $\cC$, i.e.,
$\Reg(p) = 0$ if $p \in \cC$, $\infty$ otherwise, then it can be checked that we
recover the ``generalized perceptron'' loss \cite{lecun_2005, lecun_2006} as a
special case of \eqref{eq:phi_fy_loss}. 
Proposition \ref{prop:properties_phi_fy_losses} therefore provides new
properties to understand and analyze this loss. 

Regular Fenchel-Young losses are closely related to Bregman divergences
\cite{fylosses_jmlr}.
In Appendix \ref{appendix:generalized_bregman}, we build a generalized notion of
Bregman divergence using generalized conjugates.

\paragraph{Training.}

To train regularized energy networks with our framework, the user should choose
an energy $\Phi(v, p)$, a regularization $\Omega(p)$, an output set $\cC$ (from
Proposition \ref{prop:properties_phi_fy_losses} item \ref{item:smaller_better},
the smaller this set the better)
and the model $v = g_\theta(x)$ with input $x$ and parameters $\theta$.
Given a set of input-output pairs $(x_1, y_1), \dots, (x_n, y_n) \in \cX \times
\cY$, where $\cY \subseteq \cC$, we can find the parameters $\theta$ by
minimizing the empirical risk objective regularized by $R \colon \Theta \to
\RR$,
\begin{equation}
\widehat{\theta} = \argmin_{\theta \in \Theta} 
\frac{1}{n} \sum_{i=1}^n L_\Reg^\Phi(g_\theta(x_i), y_i) + R(\theta).
\label{eq:training_objective}
\end{equation}
Thanks to the easy-to-compute gradients of $L^\Phi_\Reg$,
we can easily solve \eqref{eq:training_objective} using any (stochastic) solver. 
%In a reconstruction setting (e.g., image reconstruction)
%where there is no notion of input features, we just omit the $x_i$'s.

\section{Calibration guarantees}
\label{sec:calibration}

Many times, notably for differentiability reasons, the loss used at 
training time, here our generalized Fenchel-Young loss $L^\Phi_\Reg(v, y)$,
is used as a surrogate / proxy
for a different (potentially discrete loss) $L \colon \cY \times \cY \to \RR$,
used at test time. Calibration guarantees
\cite{zhang_2004,zhang_2004_2,bartlett_2006,steinwart_2007} ensure that
minimizing the excess of risk of the train loss will also minimize that of the
test loss (a.k.a. target loss).
We study in this section such guarantees,
assuming that $L$ satisfies an affine decomposition property
\cite{ciliberto_2016,projection_oracles}:
\begin{equation}
L(\widehat y, y) = \langle \varphi(\widehat y), V \varphi(y) + b \rangle + c(y),
\label{eq:loss_affine_decomp}
\end{equation}
where 
$\varphi(y) \mapsto V \varphi(y) + b$ is an affine map,
$\varphi(y)$ is a label embedding and
$c(y)$ is any function that depends only on $y$.
Numerous losses can be written in this form. Examples include the zero-one,
Hamming, NDCG and precision at $k$ losses \cite{nowak_2019,projection_oracles}. 
%Therefore, the loss $L$ used determines the encoding we use for $y$.
Inference in this setting works in two steps. First, we compute a ``soft''
(continuous)
prediction $p = p^\Phi_\Reg(v) \in \cC$. Second, we compute a ``hard''
(discrete)
prediction by a decoding / rounding from $\cC$ to $\cY$, 
calibrated for the loss $L$:
\begin{equation}
y_L(p) 
\coloneqq \argmin_{\widehat y \in \cY} L(\widehat y, p)
= \argmin_{\widehat y \in \cY} 
\langle \varphi(\widehat y), V \varphi(p) + b \rangle.
\label{eq:calibrated_decoding}
\end{equation}

The target risk of $f \colon \cX \to \cY$ 
and
the surrogate risk of $g \colon \cX \to \cV$ are defined by
\begin{equation}
\cL(f) \coloneqq \EE_{(X,Y) \sim \rho} ~ L(f(X), Y)
\quad \text{and} \quad
\cL^\Phi_\Reg(g) \coloneqq \EE_{(X,Y) \sim \rho} ~ L^\Phi_\Reg(g(X), Y),
\end{equation}
where $\rho$ is a typically unknown distribution over $\cX \times \cY$.
The 
%corresponding 
Bayes predictors are defined by
%\begin{equation}
$f^\star \coloneqq \argmin_{f \colon \cX \to \cY} \cL(f)$
%\quad \text{and} \quad
and
$g^\star \coloneqq \argmin_{g \colon \cX \to \cV} \cL^\Phi_\Reg(g)$.
%\end{equation}
We now establish calibration of the surrogate excess risk, under the assumption
that the energy $\Phi(v, p)$ is \textbf{linear-concave},
i.e., it can be written as 
$\Phi(v, p) = \langle v, \varphi(p) \rangle$, 
for some function $\varphi$.

\begin{proposition}{Calibration of target and surrogate excess risks}
\label{prop:calibration}

Assume $L^\Phi_\Reg(v, y)$ is $M$-smooth in $v$ w.r.t. 
the dual norm $\|\cdot\|_*$, 
$\cC$ is a compact convex set such that $\cY \subseteq \cC$
and $\Phi(v, p) = \langle v, \varphi(p) \rangle$. 
Let $\sigma \coloneqq \sup_{y \in \cY} \|V^\top \varphi(y)\|_*$.
Then, the generalized Fenchel-Young loss \eqref{eq:phi_fy_loss} is calibrated
with the target loss \eqref{eq:loss_affine_decomp} with decoder
$d = y_L \circ p^\Phi_\Reg$:
\begin{equation}
\forall g \colon \cX \to \cV \quad
\frac{(\cL(d \circ g) - \cL(f^\star))^2}{8 \sigma^2 M}
\le \cL^\Phi_\Reg(g) - \cL^\Phi_\Reg(g^\star),
\end{equation}
where $p^\Phi_\Reg \colon \cV \to \cC$ is defined in
\eqref{eq:phi_convex_argmax}
and $y_L \colon \cC \to \cY$ is defined in \eqref{eq:calibrated_decoding}.
\end{proposition}
The proof is in Appendix \ref{proof:calibration}. Proposition
\ref{prop:properties_phi_convex_conjugate} item \ref{item:smoothness} shows that
the smoothness of $\Phi$ and strong convexity of $\Reg$ ensure the smoothness of
$L_\Reg^\Phi$.
Calibration also implies Fisher consistency, namely $\cL^\Phi_\Reg(g) =
\cL^\Phi_\Reg(g^\star) \Rightarrow \cL(d \circ g) = \cL(f^\star)$,
when using the decoder $d = y_L \circ p^\Phi_\Reg$.
The existing proof technique for the calibration of regular Fenchel-Young losses
\cite{nowak_2019,projection_oracles} assumes a bilinear pairing and a loss of
the form $L_{\Reg_\varphi}(u, \varphi(y)) = \Reg_\varphi^*(u) +
\Reg_\varphi(\varphi(y)) - \langle u, \varphi(y) \rangle$, where $\Reg_\varphi$
is a strongly-convex regularizer w.r.t.\  $\varphi(y)$. Our novel proof
technique is more general, as it works with any linear-concave energy and
the regularizer $\Reg$ is w.r.t. $y$, not $\varphi(y)$. Unlike the existing
proof, our proof is valid for the pairwise model we present in the next section.

\begin{table}[t]
\caption{Multilabel classification results using various energies (test accuracy in \%).}
\centering
\begin{small}
\begin{tabular}{r c c c c c c}
\toprule
Energy & yeast & scene & mediamill & birds & emotions & cal500 \\
\midrule
Unary (linear) & 79.76 & 89.14 & 96.84 & 86.47 & 78.22 & 85.67 \\
Unary (rectifier network) & 80.03 & 91.35 & 96.91 & 91.74 & 79.79 & 86.25 \\
Pairwise & {\bf 80.19} & {\bf 91.58} & {\bf 96.95} & 91.55 & {\bf 80.56} & 85.73
\\
SPEN & 79.99 & 91.24 & 96.68 & 91.41 & 79.35 & 86.25 \\
Input-concave SPEN & 80.00 & 90.64 & {\bf 96.95} & {\bf 91.77} & 79.73 & {\bf
86.35} \\
\bottomrule
\end{tabular}
\end{small}
\label{tab:multilabel_acc}
\end{table}

\section{Experiments}
\label{sec:experiments}

\subsection{Multilabel classification}

We study in this section the application of generalized Fenchel-Young losses
to multi-label classification, setting $\cY = \{0,1\}^k$ and $\cC = [0,1]^k$, 
where $k$ is the number of labels.
When the loss $L$ in \eqref{eq:loss_affine_decomp} is the Hamming loss (1 -
accuracy), our loss is calibrated for $L$ and 
the decoding \eqref{eq:calibrated_decoding} is just
$\widehat{y}_j = 1$ if $p_j > 0.5$ else $0$.
We therefore report our empirical results using the accuracy metric.

\paragraph{Unary model.}

We consider a neural network $u = g_\theta(x) \in \RR^k$, assigning a score 
$u_j$ to each label $j \in [k]$. 
With the bilinear pairing $\Phi(u, p) = \langle u, p \rangle$, we get
\begin{equation}
p^\Phi_\Reg(u) 
%= \argmax_{p \in [0,1]^k} ~ 
%\Phi(v, p) - \Reg(p)
= \argmax_{p \in [0,1]^k} \langle u, p \rangle - \Reg(p)
= p_\Reg(u).
\label{eq:multilabel_first_order}
\end{equation}
That is, \eqref{eq:multilabel_first_order} is just a normal neural network
with $p_\Reg$ as output layer. 
When $\Reg(p) = \Reg_1(p) + \Reg_1(1 - p)$, where 
$\Reg_1(p) \coloneqq \langle p, \log p \rangle$ is Shannon's negentropy,
we get $p_\Reg(v) = \text{sigmoid}(v) \coloneqq 1 / (1 + \exp(-v))$
and
\eqref{eq:phi_fy_loss} is just the usual binary logistic / cross-entropy loss.
When $\Reg(p) = \Reg_2(p) + \Reg_2(1 - p)$, where $\Reg_2(p) \coloneqq
\frac{1}{2} \langle p, 1 - p \rangle$ is Gini's negentropy, we get a sparse
sigmoid and the binary sparsemax loss \cite[\S 6.2]{fylosses_jmlr}.
Because $\cC = [0,1]^k$ is the convex hull of $\cY = \{0,1\}^k$,
\eqref{eq:multilabel_first_order} can be intepreted as a \textbf{marginal
probability} \cite{wainwright_2008}. 
Indeed, there exists a probability distribution $\PP(Y|X)$ over $Y
\in \cY$ such that
\begin{equation}
[p_\Reg^\Phi(u)]_j = \PP(Y_j = 1 | X = x).
\end{equation}

\paragraph{Pairwise model.}

We now additionally use a network $U = h_\theta(x) \in \RR^{k \times k}$,
assigning a score $U_{i,j}$ to the pairwise interaction between labels $i$ and
$j$. With $v = (u, U)$ and the \textbf{linear-quadratic} coupling
$\Phi(v, p) = \langle u, p \rangle + \frac{1}{2} \langle p, U p \rangle$, 
we get
\begin{equation}
p^\Phi_\Reg(v) 
%= \argmax_{p \in [0,1]^k} ~ 
%\Phi(v, p) - \Reg(p)
= \argmax_{p \in [0,1]^k} ~ 
\langle u, p \rangle + 
\frac{1}{2} \langle p, U p \rangle - \Reg(p).
\label{eq:multilabel_second_order}
\end{equation}
If $U$ is negative semi-definite, 
i.e., $U = -A A^\top$ for some matrix $A \in \RR^{k \times m}$, then
the problem is concave in $p$ and can be solved optimally. 
Moreover, since $\Phi(v, p)$ is linear-concave, the calibration guarantees in
\S\ref{sec:calibration} hold and $L^\Phi_\Reg(v, y)$ is convex in $v$.
%To avoid overfitting, 
In our experiments, 
we use a matrix $A$ of rank $1$
(cf.\ Appendix \ref{appendix:experimental_details}).
Unlike \eqref{eq:closed_form_qp}, 
\eqref{eq:multilabel_second_order} does not enjoy a closed form due to the
constraints. We solve it by
coordinate ascent: if $\Reg$ is quadratic, as is the case with
Gini's negentropy, the coordinate-wise updates can be computed in closed-form.
Again,
\eqref{eq:multilabel_second_order} can be interpreted as a marginal probability.
In constrast to \eqref{eq:multilabel_second_order},
marginal inference in the closely related \textbf{Ising model}  
%\textbf{correlation polytope} \cite{wainwright_2008} 
is known to be \#P-hard~\cite{globerson_2015}. 

\paragraph{SPEN model.} 

Following SPENs \cite[Eq.\ 4 and 5]{belanger_2016},
we also tried the energy $\Phi(v, p) = \langle u, p \rangle - \Psi(w, p)$,
where $v = (u, w)$, $u = g_\theta(x)$ and $w$ are the weights of the ``prior
network'' $\Psi$ (independent of $x$). We also tried a variant where $\Psi$ is
made convex in $p$, making $\Phi(v, p)$ concave in $p$. In both cases, we
compute $p^\Phi_\Reg(v)$ by solving \eqref{eq:phi_convex_argmax} using projected
gradient ascent with backtracking linesearch.

\paragraph{Experimental setup.}

We perform experiments on 6 publicly-available datasets, see Appendix
\ref{appendix:experimental_details}.
We use the train-test split from the dataset when provided. 
When not, we use 80\% for
training data and 20\% for test data. 
For all models, we solve the outer problem \eqref{eq:training_objective} 
using ADAM. We set $\Reg(p)$ to the Gini negentropy.
We hold out 25\% of the training data for
hyperparameter validation purposes. We set $R(\theta)$ in
\eqref{eq:training_objective} to $\frac{\lambda}{2} \|\theta\|^2_2$.
For the regularization hyper-parameter $\lambda$, we search
$5$ log-spaced values between $10^{-4}$ and $10^{1}$.
For the learning rate parameter of ADAM, we search $10$ log-spaced valued
between $10^{-5}$ and $10^{-1}$.
Once we selected the best hyperparameters, we
refit the model on the entire training set.
We average results over $3$ runs with different seeds.

\paragraph{Results.}

Table \ref{tab:multilabel_acc} shows a model comparison.
We observe improvements with the pairwise model on $4$ out of $6$
datasets, confirming that our losses are able to learn useful models. 
Input-concavity helps improve SPENs in $4$ out of $6$ datasets.
Table \ref{tab:multilabel_grad} 
confirms that using the envelope theorem for computing gradients works
comparably to (if not better than) the implicit function theorem.
Table \ref{tab:multilabel_losses} shows that our losses 
outperform the energy, the cross-entropy and the generalized
perceptron losses.

\vspace{-0.3cm}
\subsection{Imitation learning}

In this section, we study the application of generalized Fenchel-Young losses to
imitation learning. This setting consists in learning a policy
$\pi: \cX \mapsto \cY$, a mapping from states $\cX$ to actions $\cY$, from
a fixed dataset of expert demonstrations $(x_1, y_1), \dots, (x_n, y_n) \in \cX
\times \cY$. In particular, we only consider a Behavior Cloning approach
\cite{pomerleau1991efficient}, which essentially reduces imitation learning to
supervised learning (as opposed to inverse RL methods
\cite{russell1998learning, ng2000algorithms}). The learned policy $\pi$ is
evaluated on its performance, which is the expected sum of rewards of
the environment \cite{sutton2018reinforcement}.

\paragraph{Experimental setup.} 

We consider four MuJoCo Gym locomotion environments \cite{openaigym} together
with the demonstrations provided by Orsini et al.~\cite{orsini2021matters};
see Appendix \ref{appendix:imitation_learning}
%for environment and demonstration details. 
for details. 
We evaluate the learned policy
$\pi$ for different number of demonstration trajectories: 1, 4 and 11,
consistently with
\cite{ho2016generative,kostrikov2018discriminator,dadashi2020primal}.  The
action space in the demonstrations is included in $[-1, 1]^k$, where the
dimensionality of the action space $k$ corresponds to the torques of the
actuators. We scale the action space to the hypercube $\cY = \cC = [0, 1]^k$ at
learning time and scale it back to the original action space at inference time.
Similarly to the multilabel classification setup, we evaluate the unary
and pairwise models, with the only difference that we use two hidden layers
instead of one, since it leads to significantly better performance. 
We specify the hyperparameter selection procedure in
Appendix \ref{appendix:imitation_learning}.

\paragraph{Results.}

\begin{figure}[t]
\centering
\includegraphics[width=0.80\linewidth]{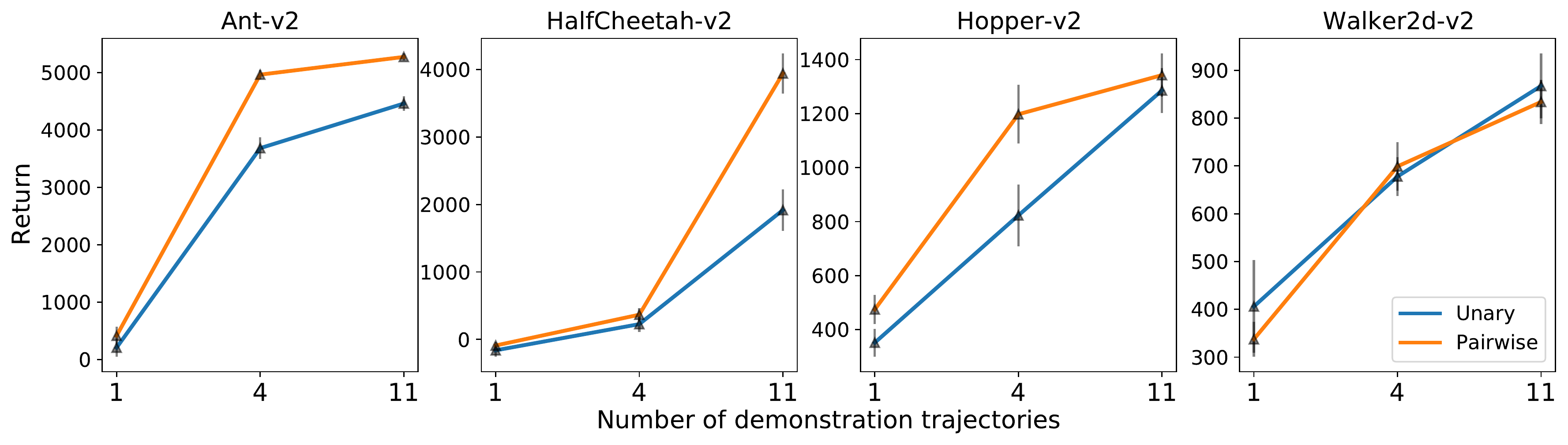}
\caption{Average performance (higher is better) and standard deviation over 10
seeds.}
\label{figure:imitation_learning_results}
\end{figure}

We run the best hyperparameters over
10 seeds and report final performance over 100 evaluation episodes. Figure
\ref{figure:imitation_learning_results} shows a clear improvement of the
pairwise model over the unary model for 3 out of 4 tasks.
Contrary to the unary model, the pairwise model enables to
capture the interdependence between the different
torques of the action space, translating into better performance. 

\vspace{-0.4cm}
\section{Conclusion}

Building upon generalized conjugate functions, 
we proposed generalized Fenchel-Young
losses, a natural loss construction for learning energy networks and studied its
properties. Thanks to
conditions on the energy $\Phi$ and the regularizer $\Reg$ ensuring the
loss smoothness, we established calibration guarantees for the case of
linear-concave energies, a more general result than the existing analysis,
restricted to bilinear energies. We demonstrated the effectiveness of our losses
on multilabel classification and imitation learning tasks. We hope that this
paper will help popularize generalized conjugates as a powerful tool for machine
learning and optimization.

\newpage

\section*{Acknowledgments}

MB thanks Gabriel Peyr\'{e} for numerous discussions on $C$-transforms
and
Clarice Poon for discussions on envelope theorems,
as well as
Vlad Niculae and Andr\'{e} Martins for many fruitful discussions.

%\bibliography{paper}

\begin{thebibliography}{10}

\bibitem{ablin_2020}
P.~Ablin, G.~Peyr{\'e}, and T.~Moreau.
\newblock Super-efficiency of automatic differentiation for functions defined
  as a minimum.
\newblock In {\em Proc. of ICML}, pages 32--41. PMLR, 2020.

\bibitem{acharyya_2013}
S.~Acharyya.
\newblock Learning to rank in supervised and unsupervised settings using
  convexity and monotonicity.
\newblock 2013.

\bibitem{amari_2016}
S.~Amari.
\newblock {\em
  \href{https://www.springer.com/us/book/9784431559771}{Information Geometry
  and Its Applications}}.
\newblock Springer, 2016.

\bibitem{icnn_icml}
B.~Amos, L.~Xu, and J.~Z. Kolter.
\newblock Input convex neural networks.
\newblock In {\em ICML}, pages 146--155. PMLR, 2017.

\bibitem{aubin_2022}
P.-C. Aubin-Frankowski and S.~Gaubert.
\newblock The tropical analogues of reproducing kernels.
\newblock {\em arXiv preprint arXiv:2202.11410}, 2022.

\bibitem{bakir_2007}
G.~BakIr, T.~Hofmann, B.~Sch{\"o}lkopf, A.~J. Smola, and B.~Taskar.
\newblock {\em Predicting structured data}.
\newblock MIT press, 2007.

\bibitem{bregman_clustering}
A.~Banerjee, S.~Merugu, I.~S. Dhillon, J.~Ghosh, and J.~Lafferty.
\newblock Clustering with bregman divergences.
\newblock {\em Journal of machine learning research}, 6(10), 2005.

\bibitem{exponential_families}
O.~Barndorff-Nielsen.
\newblock {\em
  \href{https://onlinelibrary.wiley.com/doi/book/10.1002/9781118857281}{Information
  and Exponential Families: In Statistical Theory}}.
\newblock John Wiley \& Sons, 1978.

\bibitem{bartlett_2006}
P.~L. Bartlett, M.~I. Jordan, and J.~D. McAuliffe.
\newblock Convexity, classification, and risk bounds.
\newblock {\em Journal of the American Statistical Association},
  101(473):138--156, 2006.

\bibitem{beck_2017}
A.~Beck.
\newblock {\em First-order methods in optimization}.
\newblock SIAM, 2017.

\bibitem{belanger_2016}
D.~Belanger and A.~McCallum.
\newblock Structured prediction energy networks.
\newblock In {\em International Conference on Machine Learning}, pages
  983--992. PMLR, 2016.

\bibitem{belanger_2013}
D.~Belanger, D.~Sheldon, and A.~McCallum.
\newblock
  \href{http://www.cmap.polytechnique.fr/~jaggi/NeurIPS-workshop-FW-greedy/papers/belanger_sheldon_mccallum_final.pdf}{Marginal
  inference in {MRF}s using {F}rank-{W}olfe}.
\newblock In {\em NeurIPS Workshop on Greedy Opt., FW and Friends}, 2013.

\bibitem{belanger_2017}
D.~Belanger, B.~Yang, and A.~McCallum.
\newblock End-to-end learning for structured prediction energy networks.
\newblock In {\em International Conference on Machine Learning}, pages
  429--439. PMLR, 2017.

\bibitem{bell_2008}
B.~M. Bell and J.~V. Burke.
\newblock Algorithmic differentiation of implicit functions and optimal values.
\newblock In {\em Advances in Automatic Differentiation}, pages 67--77.
  Springer, 2008.

\bibitem{bertsekas_1997}
D.~P. Bertsekas.
\newblock Nonlinear programming.
\newblock {\em Journal of the Operational Research Society}, 48(3):334--334,
  1997.

\bibitem{projection_oracles}
M.~Blondel.
\newblock Structured prediction with projection oracles.
\newblock {\em Advances in neural information processing systems}, 32, 2019.

\bibitem{blondel_implicit_diff}
M.~Blondel, Q.~Berthet, M.~Cuturi, R.~Frostig, S.~Hoyer, F.~Llinares-L{\'o}pez,
  F.~Pedregosa, and J.-P. Vert.
\newblock Efficient and modular implicit differentiation.
\newblock {\em arXiv preprint arXiv:2105.15183}, 2021.

\bibitem{fylosses_jmlr}
M.~Blondel, A.~F. Martins, and V.~Niculae.
\newblock Learning with fenchel-young losses.
\newblock {\em JMLR}, 21(35):1--69, 2020.

\bibitem{bonnans_2013}
J.~F. Bonnans and A.~Shapiro.
\newblock {\em Perturbation analysis of optimization problems}.
\newblock Springer Science \& Business Media, 2013.

\bibitem{borwein_2006}
J.~Borwein and A.~Lewis.
\newblock {\em Convex Analysis}.
\newblock 2006.

\bibitem{bottou_2018}
L.~Bottou, M.~Arjovsky, D.~Lopez-Paz, and M.~Oquab.
\newblock Geometrical insights for implicit generative modeling.
\newblock pages 229--268. 2018.

\bibitem{boyd_book}
S.~Boyd, S.~P. Boyd, and L.~Vandenberghe.
\newblock {\em Convex optimization}.
\newblock Cambridge university press, 2004.

\bibitem{openaigym}
G.~Brockman, V.~Cheung, L.~Pettersson, J.~Schneider, J.~Schulman, J.~Tang, and
  W.~Zaremba.
\newblock Openai gym, 2016.

\bibitem{calafiore_2019}
G.~C. Calafiore, S.~Gaubert, and C.~Possieri.
\newblock Log-sum-exp neural networks and posynomial models for convex and
  log-log-convex data.
\newblock {\em IEEE transactions on neural networks and learning systems},
  31(3):827--838, 2019.

\bibitem{chancelier_2018}
J.-P. Chancelier and M.~De~Lara.
\newblock Fenchel-moreau conjugation inequalities with three couplings and
  application to stochastic bellman equation.
\newblock {\em arXiv preprint arXiv:1804.03034}, 2018.

\bibitem{ciliberto_2016}
C.~Ciliberto, L.~Rosasco, and A.~Rudi.
\newblock A consistent regularization approach for structured prediction.
\newblock In {\em Proc. of NeurIPS}, pages 4412--4420, 2016.

\bibitem{structured_perceptron}
M.~Collins.
\newblock \href{https://dl.acm.org/citation.cfm?id=1118694}{Discriminative
  training methods for Hidden Markov Models: Theory and experiments with
  perceptron algorithms}.
\newblock In {\em Proc. of EMNLP}, 2002.

\bibitem{dadashi2020primal}
R.~Dadashi, L.~Hussenot, M.~Geist, and O.~Pietquin.
\newblock Primal wasserstein imitation learning.
\newblock {\em International Conference on Learning Representations}, 2021.

\bibitem{danskin_1967}
J.~M. Danskin.
\newblock {\em The theory of max-min and its application to weapons allocation
  problems}.
\newblock Springer Science \& Business Media, 1967.

\bibitem{duchi_2016}
J.~C. Duchi, K.~Khosravi, and F.~Ruan.
\newblock \href{http://arxiv.org/abs/1603.00126}{Multiclass classification,
  information, divergence, and surrogate risk}.
\newblock {\em The Annals of Statistics}, 46(6B):3246--3275, 2018.

\bibitem{ghadimi_2018}
S.~Ghadimi and M.~Wang.
\newblock Approximation methods for bilevel programming.
\newblock {\em arXiv preprint arXiv:1802.02246}, 2018.

\bibitem{globerson_2015}
A.~Globerson, T.~Roughgarden, D.~Sontag, and C.~Yildirim.
\newblock How hard is inference for structured prediction?
\newblock In {\em International Conference on Machine Learning}, pages
  2181--2190. PMLR, 2015.

\bibitem{glorot_2011}
X.~Glorot, A.~Bordes, and Y.~Bengio.
\newblock Deep sparse rectifier neural networks.
\newblock In {\em AISTATS}, pages 315--323. JMLR Workshop and Conference
  Proceedings, 2011.

\bibitem{maxout}
I.~Goodfellow, D.~Warde-Farley, M.~Mirza, A.~Courville, and Y.~Bengio.
\newblock Maxout networks.
\newblock In {\em ICML}, pages 1319--1327. PMLR, 2013.

\bibitem{no_mcmc}
W.~Grathwohl, J.~Kelly, M.~Hashemi, M.~Norouzi, K.~Swersky, and D.~Duvenaud.
\newblock No mcmc for me: Amortized samplers for fast and stable training of
  energy-based models.
\newblock 2021.

\bibitem{griewank_2008}
A.~Griewank and A.~Walther.
\newblock {\em Evaluating derivatives: principles and techniques of algorithmic
  differentiation}.
\newblock SIAM, 2008.

\bibitem{hiriart_1993}
J.-B. Hiriart-Urruty and C.~Lemar{\'e}chal.
\newblock {\em Convex analysis and minimization algorithms II}, volume 305.
\newblock Springer science \& business media, 1993.

\bibitem{ho2016generative}
J.~Ho and S.~Ermon.
\newblock Generative adversarial imitation learning.
\newblock In {\em Advances in Neural Information Processing Systems}, 2016.

\bibitem{kakade_2009}
S.~Kakade, S.~Shalev-Shwartz, A.~Tewari, et~al.
\newblock On the duality of strong convexity and strong smoothness: Learning
  applications and matrix regularization.
\newblock {\em Tech report}, 2(1):35, 2009.

\bibitem{kostrikov2018discriminator}
I.~Kostrikov, K.~K. Agrawal, D.~Dwibedi, S.~Levine, and J.~Tompson.
\newblock Discriminator-actor-critic: Addressing sample inefficiency and reward
  bias in adversarial imitation learning.
\newblock {\em International Conference on Learning Representations}, 2019.

\bibitem{krantz_2012}
S.~G. Krantz and H.~R. Parks.
\newblock {\em The implicit function theorem: history, theory, and
  applications}.
\newblock Springer Science \& Business Media, 2012.

\bibitem{barrierfw}
R.~G. Krishnan, S.~Lacoste-Julien, and D.~Sontag.
\newblock \href{https://arxiv.org/abs/1511.02124}{Barrier Frank-Wolfe for
  marginal inference}.
\newblock In {\em Proc. of NeurIPS}, 2015.

\bibitem{Lafferty2001}
J.~D. Lafferty, A.~McCallum, and F.~C. Pereira.
\newblock \href{http://dl.acm.org/citation.cfm?id=645530.655813}{Conditional
  Random Fields: Probabilistic models for segmenting and labeling sequence
  data}.
\newblock In {\em Proc. of ICML}, 2001.

\bibitem{lecun_2006}
Y.~LeCun, S.~Chopra, R.~Hadsell, M.~Ranzato, and F.~Huang.
\newblock A tutorial on energy-based learning.
\newblock {\em Predicting structured data}, 1(0), 2006.

\bibitem{lecun_2005}
Y.~LeCun and F.~J. Huang.
\newblock Loss functions for discriminative training of energy-based models.
\newblock In {\em International Workshop on Artificial Intelligence and
  Statistics}, pages 206--213. PMLR, 2005.

\bibitem{lin_2020}
T.~Lin, C.~Jin, and M.~Jordan.
\newblock On gradient descent ascent for nonconvex-concave minimax problems.
\newblock In {\em International Conference on Machine Learning}, pages
  6083--6093. PMLR, 2020.

\bibitem{sparsemax}
A.~F. Martins and R.~F. Astudillo.
\newblock \href{https://arxiv.org/abs/1602.02068} {From softmax to sparsemax: A
  sparse model of attention and multi-label classification}.
\newblock In {\em Proc. of ICML}, 2016.

\bibitem{mccullagh_1989}
P.~McCullagh and J.~A. Nelder.
\newblock {\em
  \href{https://www.crcpress.com/Generalized-Linear-Models/McCullagh-Nelder/p/book/9780412317606}{Generalized
  Linear Models}}, volume~37.
\newblock CRC press, 1989.

\bibitem{milgrom_2002}
P.~Milgrom and I.~Segal.
\newblock Envelope theorems for arbitrary choice sets.
\newblock {\em Econometrica}, 70(2):583--601, 2002.

\bibitem{moreau_1966}
J.-J. Moreau.
\newblock Fonctionnelles convexes.
\newblock {\em S{\'e}minaire Jean Leray}, (2):1--108, 1966.

\bibitem{nash_2019}
C.~Nash and C.~Durkan.
\newblock Autoregressive energy machines.
\newblock In {\em International Conference on Machine Learning}, pages
  1735--1744. PMLR, 2019.

\bibitem{glm}
J.~A. Nelder and R.~J. Baker.
\newblock {\em
  \href{https://onlinelibrary.wiley.com/doi/full/10.1002/0471667196.ess0866.pub2}{Generalized
  Linear Models}}.
\newblock Wiley Online Library, 1972.

\bibitem{nesterov_2003}
Y.~Nesterov.
\newblock {\em Introductory lectures on convex optimization: A basic course},
  volume~87.
\newblock Springer Science \& Business Media, 2003.

\bibitem{ng2000algorithms}
A.~Y. Ng, S.~J. Russell, et~al.
\newblock Algorithms for inverse reinforcement learning.
\newblock In {\em International Conference on Machine Learning}, 2000.

\bibitem{sparsemap}
V.~Niculae, A.~F. Martins, M.~Blondel, and C.~Cardie.
\newblock \href{https://arxiv.org/abs/1802.04223}{SparseMAP: Differentiable
  sparse structured inference}.
\newblock In {\em Proc. of ICML}, 2018.

\bibitem{nowak_2019}
A.~Nowak-Vila, F.~Bach, and A.~Rudi.
\newblock A general theory for structured prediction with smooth convex
  surrogates.
\newblock {\em arXiv preprint arXiv:1902.01958}, 2019.

\bibitem{orabona_2019}
F.~Orabona.
\newblock A modern introduction to online learning.
\newblock {\em arXiv preprint arXiv:1912.13213}, 2019.

\bibitem{orsini2021matters}
M.~Orsini, A.~Raichuk, L.~Hussenot, D.~Vincent, R.~Dadashi, S.~Girgin,
  M.~Geist, O.~Bachem, O.~Pietquin, and M.~Andrychowicz.
\newblock What matters for adversarial imitation learning?
\newblock {\em Advances in Neural Information Processing Systems}, 2021.

\bibitem{osokin_2017}
A.~Osokin, F.~Bach, and S.~Lacoste-Julien.
\newblock On structured prediction theory with calibrated convex surrogate
  losses.
\newblock In {\em Proc. of NIPS}, pages 302--313, 2017.

\bibitem{peyre_2019}
G.~Peyr{\'e}, M.~Cuturi, et~al.
\newblock Computational optimal transport: With applications to data science.
\newblock {\em Foundations and Trends{\textregistered} in Machine Learning},
  11(5-6):355--607, 2019.

\bibitem{pomerleau1991efficient}
D.~A. Pomerleau.
\newblock Efficient training of artificial neural networks for autonomous
  navigation.
\newblock {\em Neural computation}, 1991.

\bibitem{rockafellar_2009}
R.~T. Rockafellar and R.~J.-B. Wets.
\newblock {\em Variational analysis}, volume 317.
\newblock Springer Science \& Business Media, 2009.

\bibitem{rubinov_2000}
A.~M. Rubinov.
\newblock {\em Abstract convexity and global optimization}.
\newblock Springer Science \& Business Media, 2000.

\bibitem{russell1998learning}
S.~Russell.
\newblock Learning agents for uncertain environments.
\newblock In {\em Conference on Computational learning theory}, 1998.

\bibitem{santambrogio_2015}
F.~Santambrogio.
\newblock Optimal transport for applied mathematicians.
\newblock {\em Birk{\"a}user, NY}, 55(58-63):94, 2015.

\bibitem{singer_1997}
I.~Singer.
\newblock {\em Abstract convex analysis}, volume~25.
\newblock John Wiley \& Sons, 1997.

\bibitem{song_2021}
Y.~Song and D.~P. Kingma.
\newblock How to train your energy-based models.
\newblock {\em arXiv preprint arXiv:2101.03288}, 2021.

\bibitem{steinwart_2007}
I.~Steinwart.
\newblock How to compare different loss functions and their risks.
\newblock {\em Constructive Approximation}, 26(2):225--287, 2007.

\bibitem{sutton_introduction_2011}
C.~Sutton, A.~McCallum, et~al.
\newblock An introduction to conditional random fields.
\newblock {\em Foundations and Trends in Machine Learning}, 4(4):267--373,
  2012.

\bibitem{sutton2018reinforcement}
R.~S. Sutton and A.~G. Barto.
\newblock {\em Reinforcement learning: An introduction}.
\newblock 2018.

\bibitem{wainwright_2008}
M.~J. Wainwright, M.~I. Jordan, et~al.
\newblock Graphical models, exponential families, and variational inference.
\newblock {\em Foundations and Trends{\textregistered} in Machine Learning},
  1(1--2):1--305, 2008.

\bibitem{yuille_2003}
A.~L. Yuille and A.~Rangarajan.
\newblock The concave-convex procedure.
\newblock {\em Neural computation}, 15(4):915--936, 2003.

\bibitem{zhang_2004}
T.~Zhang.
\newblock Statistical analysis of some multi-category large margin
  classification methods.
\newblock {\em Journal of Machine Learning Research}, 5(Oct):1225--1251, 2004.

\bibitem{zhang_2004_2}
T.~Zhang.
\newblock Statistical behavior and consistency of classification methods based
  on convex risk minimization.
\newblock {\em The Annals of Statistics}, 32(1):56--85, 2004.

\bibitem{zhou_2018}
X.~Zhou.
\newblock On the fenchel duality between strong convexity and lipschitz
  continuous gradient.
\newblock {\em arXiv preprint arXiv:1803.06573}, 2018.

\end{thebibliography}
%\bibliographystyle{abbrv}

\clearpage

\appendix

\section{Generalized Bregman divergences}
\label{appendix:generalized_bregman}

The expression $\Reg^*(v) + \Reg(p) - \langle v, p \rangle$ at the heart of
Fenchel-Young losses is closely related to Bregman divergences; see
\cite[Theorem 1.1]{amari_2016} and \cite{fylosses_jmlr}.
In this section, we develop a similar relationship between generalized
Fenchel-Young losses and a new generalized notion of Bregman divergence.

\paragraph{Biconjugates.}

We begin by recalling well-known results on biconjugates \cite{boyd_book}.
Applying the conjugate \eqref{eq:convex_conjugate} twice, we obtain the
biconjugate $\Reg^{**}(p)$. It is well-known that a function $\Reg$ 
is convex
\textit{and}
closed (i.e., lower-semicontinuous) if and only if $\Reg = \Reg^{**}$. 
This therefore provides a variational characterization of lower-semicontinuous
convex functions. This characterization naturally motivates the class of
$\Phi$-convex functions (\S\ref{sec:phi_convex_conjugates}).
If $\Reg$ is nonconvex, $\Reg^{**}$ is $\Reg$'s 
tightest convex lower bound.

\paragraph{Generalized biconjugates.}

The generalized conjugate in \eqref{eq:phi_convex_conjugate} uses maximization 
w.r.t.\ the second argument $p \in \cC$. To obtain a generalized conjugate whose
maximization is w.r.t.\ the first argument $v \in \cV$ instead, we define
$\bPhi(p, v) \coloneqq \Phi(v, p)$.
Note that if $\Phi$ is symmetric, the distinction between $\Phi$ and $\bPhi$ is
not necessary.
Similarly to \eqref{eq:phi_convex_conjugate}, we can then define
the $\bPhi$-conjugate of $\Lambda \colon \cV \to \RR$ as
\begin{equation}
\Lambda^\bPhi(p) 
= \max_{v \in \cV} ~ \bPhi(p, v) - \Lambda(v)
= \max_{v \in \cV} ~ \Phi(v, p) - \Lambda(v),
\label{eq:theta_max}
\end{equation}
i.e., the maximization is over the left argument of $\Phi$.
We define the corresponding argmax as
\begin{equation}
v_\Lambda^\bPhi(p) \coloneqq 
\argmax_{v \in \cV} ~ \bPhi(p, v) - \Lambda(v)
= \argmax_{v \in \cV} ~ \Phi(v, p) - \Lambda(v).
\label{eq:theta_argmax}
\end{equation}
In particular, with $\Lambda \coloneqq \Reg^\Phi$, 
we can define the generalized biconjugate 
$\Lambda^\bPhi = \Reg^{\Phi\bPhi} \colon \cC \to \RR$.
Generalized biconjugates enjoy similar properties as regular biconjugates, 
as we now show.
\begin{proposition}[Properties of generalized biconjugates]

Let $\Reg \colon \cC \to \RR$,
$\Phi \colon \cV \times \cC \to \RR$
and $\Psi(p, v) \coloneqq \Phi(v, p)$.
\begin{enumerate}[topsep=0pt,itemsep=2pt,parsep=2pt,leftmargin=10pt]

\item \normalfont{\textbf{Lower-bound:}} 
$\Reg^{\Phi\bPhi}(p) \le \Reg(p)$ for all $p \in \cC$.

\item \normalfont{\textbf{Equality:}}
$\Reg^{\Phi\bPhi}(p) = \Reg(p)$ if and only if $\Reg$ is $\Psi$-convex.

\item \normalfont{\textbf{Tightest lower-bound:}} 
$\Reg^{\Phi\bPhi}(p)$ is the tightest $\Psi$-convex lower-bound of $\Reg(p)$.

\end{enumerate}
\label{prop:properties_phi_biconjugates}
\end{proposition}
Proofs are given in Appendix \ref{proof:properties_phi_biconjugates}.
Similar results hold for $\Lambda^{\Psi\Phi}$, where $\Lambda \colon \cV \to
\RR$.

\paragraph{Definition.}

We can now define the \textbf{generalized Bregman divergence}
$D_\Reg^\Phi \colon \cC \times \cC \to \RR_+$ as 
\begin{equation}
D_\Reg^\Phi(p, p')
\coloneqq \Reg(p) - \Phi(v_{\Reg^\Phi}^\bPhi(p'), p)
- \Reg(p') + \Phi(v_{\Reg^\Phi}^\bPhi(p'), p'),
\label{eq:phi_bregman_div}
\end{equation}
where we recall that $\bPhi(p, v) \coloneqq \Phi(v, p)$
and using \eqref{eq:theta_argmax} we have
\begin{equation}
v_{\Reg^\Phi}^\bPhi(p)
= \argmax_{v \in \cV} ~ \bPhi(p, v) - \Reg^\Phi(v)
= \argmax_{v \in \cV} ~ \Phi(v, p) - \Reg^\Phi(v).
\label{eq:theta_argmax_f_phi}
\end{equation}
We have therefore obtained a notion of Bregman divergence parametrized by a
coupling $\Phi(v, p)$, such as a neural network.

As shown in the proposition below, 
generalized Bregman divergences enjoy similar properties as regular Bregman
divergences.
Proofs are given in Appendix \ref{proof:properties_Bregman}.
\newpage

\begin{proposition}[Properties of generalized Bregman divergences]
Let $\Reg$ be a $\Psi$-convex function,
where $\Psi(p, v) \coloneqq \Phi(v, p)$.
\begin{enumerate}[topsep=0pt,itemsep=2pt,parsep=2pt,leftmargin=10pt]

\item \normalfont{\textbf{Link with generalized FY loss:}}
Denoting $v \coloneqq v_{\Reg^\Phi}^\bPhi(p')$, 
we have $D_\Reg^\Phi(p, p') = L_\Reg^\Phi(v, p)$.

\item \normalfont{\textbf{Non-negativity:}}
$D_\Reg^\Phi(p, p') \ge 0$, for all $p, p' \in \cC$.

\item \normalfont{\textbf{Identity of indiscernibles:}}
$D_\Reg^\Phi(p, p') = 0 \Leftrightarrow p = p'$ if $p \mapsto \Phi(v, p) -
\Reg(p)$ is strictly concave for all $v \in \cV$.

\item \normalfont{\textbf{Convexity:}}
    if $p \mapsto \Phi(v, p) - \Reg(p)$ is concave for all $v \in \cV$,
then $D_\Reg^\Phi(p, p')$ is convex in $p$.

\item \normalfont{\textbf{Recovering Bregman divergences:}}
If $\Phi(v, p)$ is the bilinear
coupling $\langle v, p \rangle$, 
%which presupposes that $\cV$ and $\cC$ have compatible dimensions, 
then we recover
the usual Bregman divergence
$D_\Reg \colon \cC \times \cC \to \RR_+$
\begin{equation}
D_\Reg^\Phi(p, p') = 
D_\Reg(p, p') \coloneqq 
\Reg(p) - \Reg(p') - \langle \nabla \Reg(p'), p - p' \rangle.
\label{eq:bregman_div}
\end{equation}

\end{enumerate}
\label{prop:properties_Bregman}
\end{proposition}
Some remarks:
\begin{itemize}[topsep=0pt,itemsep=2pt,parsep=2pt,leftmargin=10pt]

\item The generalized Bregman divergence is between objects $p$ and $p'$ of the
    same space $\cC$, while the generalized Fenchel-Young loss is between
    objects $v$ and $p$ of mixed spaces $\cV$ and $\cC$.

\item If $\Phi(v, p) - \Reg(p)$ is concave in $p$, then $D_\Reg^\Phi(p, p')$ is 
convex in $p$, as is the case of the usual Bregman divergence $D_\Reg(p, p')$. 
However, \eqref{eq:theta_argmax_f_phi} is not easy to solve 
globally in general, as it is the maximum of a difference of convex functions in
$v$.  This can be done approximately by using the convex-concave procedure
\cite{yuille_2003},
linearizing the left part. We have the opposite situation with the
generalized Fenchel-Young loss: if $\Phi(v, p) - \Reg(p)$ is convex-concave,
$L_\Reg^\Phi(v, y)$ is easy to compute but it is a difference of convex
functions in $v$.

\item Similarly, we may also define
$D_{\Reg^\Phi}^\bPhi(v, v')
= \Reg^\Phi(v) - \bPhi(p_\Reg^\Phi(v'), v)
- \Reg^\Phi(v') + \bPhi(p_\Reg^\Phi(v'), v')$,
where we recall that
$p^\Phi_\Reg(v) \coloneqq \argmax_{p \in \cC} \Phi(v, p) - \Reg(p)$.
We have thus obtained a divergence between two objects $v$ and $v'$ in
the same space $\cV$.
\end{itemize}

\section{Proofs}

\subsection{Proofs for Proposition \ref{prop:properties_phi_convex_conjugate}
(properties of generalized conjugates)}
\label{proof:properties_phi_convex_conjugate}

\begin{enumerate}[topsep=0pt,itemsep=2pt,parsep=2pt,leftmargin=10pt]

\item \textbf{Generalized Fenchel-Young inequality.} 
From \eqref{eq:phi_convex_conjugate}, we immediately obtain 
$\Reg^\Phi(v) \ge \Phi(v, p) - \Reg(p)$ 
for all $v \in \cV$ 
and all $p \in \cC$.

\item \textbf{Convexity.} 
If $\Phi(v, p)$ is convex in $v$, 
then $\Reg^\Phi(v)$ is the maximum of a
family of convex functions indexed by $p$. 
Therefore, $\Reg^\Phi(v)$ is convex.
Note that this is the case even if $\Reg(p)$ is nonconvex.

\item \textbf{Order reversing.} Since $\Reg \le \Lambda$, we have
\begin{equation}
\Reg^\Phi(v) 
= \max_{p \in \cC} \Phi(v, p) - \Reg(p)
\ge \max_{p \in \cC} \Phi(v, p) - \Lambda(p)
= \Lambda^\Phi(v).
\end{equation}

\item \textbf{Continuity.} See \cite[Box 1.8]{santambrogio_2015}.

\item \textbf{Gradient.} 
See ``Assumptions for envelope theorems'' in \S\ref{sec:phi_convex_conjugates}.

\item \textbf{Smoothness.} 
We follow the proof technique of \cite[Lemma 4.3]{lin_2020}, which states that
$v \mapsto \max_{p \in \cC} E(v, p)$ is $(\beta+\beta^2/\gamma)$-smooth if $E$ is
$\beta$-smooth in $(v, p)$ and $\gamma$-strongly concave in $p$ over $\cC$. We
show here that if $E(v, p)$ decomposes as
$E(v, p) = \Phi(v, p) - \Reg(p)$, where $\Reg$ is
$\gamma$-strongly convex over $\cC$ \textbf{but possibly nonsmooth}
(as is the case for instance of Shannon's negentropy), 
and $\Phi(v, p)$ is $\beta$-smooth in $(v,p)$ and concave in $p$, then
we still have that
$\Reg^\Phi(v) = \max_{p \in \cC} \Phi(v, p) - \Reg(p)$ is
$(\beta+\beta^2/\gamma)$-smooth. 

For brevity, let us define the shorthand $p^\Phi_\Reg(v) \coloneqq p^\star(v)$.
From \cite[Section A.3]{lin_2020}, we have
\begin{equation}
\gamma \|p^\star(v_2) - p^\star(v_1)\|^2 
\le
\langle p^\star(v_2) - p^\star(v_1),
\nabla_2 E(v_2, p^\star(v_2)) - \nabla_2 E(v_1, p^\star(v_2)) \rangle
\end{equation}
for all $v_1, v_2 \in \cV$.
With $E(v, p) = \Phi(v, p) - \Reg(p)$, we obtain
\begin{equation}
\gamma \|p^\star(v_2) - p^\star(v_1)\|^2 
\le
\langle p^\star(v_2) - p^\star(v_1),
\nabla_2 \Phi(v_2, p^\star(v_2)) - \nabla_2 \Phi(v_1, p^\star(v_2)) \rangle.
\end{equation}
Since $\Phi$ is $\beta$-smooth, we have for all $p \in \cC$
\begin{equation}
\|\nabla_2 \Phi(v_2, p) - \nabla_2 \Phi(v_1, p)\|_* 
\le 
\beta \|(v_2, p) - (v_1, p)\|
= \beta \|v_2 - v_1\|.
\end{equation}
Combined with the Holder inequality $\langle a, b \rangle \le \|a\|_* \|b\|$,
this gives
\begin{equation}
\gamma \|p^\star(v_2) - p^\star(v_1)\|^2 
\le
\beta \|p^\star(v_2) - p^\star(v_1)\| \|v_2 - v_1\|.
\end{equation}
Simplifying, we get
\begin{equation}
\|p^\star(v_2) - p^\star(v_1)\|
\le
\frac{\beta}{\gamma} \|v_2 - v_1\|.
\label{eq:p_star_lipschitz}
\end{equation}
Therefore, $p^\Phi_\Reg(v) = p^\star(v)$ is $\beta/\gamma$-Lipschitz.

From Rockafellar's envelope theorem \cite[Theorem 10.31]{rockafellar_2009},
$\nabla \Reg^\Phi(v) = \nabla_1 \Phi(v, p^\star(v))$. Since $\Phi$ is
$\beta$-smooth, we therefore have
\begin{align}
\|\nabla \Reg^\Phi(v_2) - \nabla \Reg^\Phi(v_1)\|_*
&= \|\nabla_1 \Phi(v_2, p^\star(v_2)) - \nabla_1 \Phi(v_1, p^\star(v_1))\|_* \\
&\le \beta \|(v_2 - p^\star(v_2)) - (v_1, p^\star(v_1))\| \\
&\le \beta (\|v_2 - v_1\| + \|p^\star(v_2) - p^\star(v_1)\|) \\
&\le \left(\beta + \frac{\beta^2}{\gamma}\right) \|v_2 - v_1\| \\
&\le 2 \frac{\beta^2}{\gamma} \|v_2 - v_1\|,
\end{align}
where we used \eqref{eq:p_star_lipschitz} and $\frac{\beta}{\gamma} \ge 1$.

A related result in the context of bilevel programming but with
different assumptions and proof is stated in \cite[Lemma 2.2]{ghadimi_2018}.
More precisely, the proof of that result requires twice differentiability of
$E(v, p) = \Phi(v, p) - \Reg(p)$ while we do not. 
Moreover, applying that result to our setting would
require $E(v, p)$ to be smooth in $(v,p)$ while we only assume $\Phi(v, p)$ to
be the case. We emphasize again that $\Reg(p)$ is nonsmooth when it is the
negentropy.

\end{enumerate}

\subsection{Proofs for Proposition \ref{prop:closed_forms} (closed forms)}
\label{proof:closed_forms}

\begin{enumerate}[topsep=0pt,itemsep=2pt,parsep=2pt,leftmargin=10pt]

\item \textbf{Bilinear coupling.}
This follows from
\begin{equation}
\Reg^\Phi(v) 
= \max_{p \in \cC} \langle v, U p \rangle - \Reg(p)
= \max_{p \in \cC} \langle U^\top v, p \rangle - \Reg(p)
= \Reg^*(U^\top v)
\end{equation}
and similarly for $p_\Reg(v)$.

\item \textbf{Linear-quadratic coupling.} Let us define the function
\begin{equation}
F(p) = 
\frac{\gamma}{2} \|p\|^2_2 
- \frac{1}{2} \langle p, A p \rangle 
- \langle p, b \rangle
= \frac{1}{2} \langle p, (\gamma I - A) p \rangle
- \langle p, b \rangle.
\end{equation}
Its gradient is $\nabla F(p) = (\gamma I - A)p - b$.
Setting $\nabla F(p^\star) = 0$, we obtain
\begin{equation}
(\gamma I - A)p^\star = b
\Leftrightarrow 
p^\star = (\gamma I - A)^{-1} b,
\end{equation}
where we assumed that $(\gamma I - A)$ is positive definite.
We therefore get
\begin{equation}
F(p^\star) = 
\frac{1}{2} \langle p^\star, b \rangle
- \langle p^\star, b \rangle
= -\frac{1}{2} \langle p^\star, b \rangle
= -\frac{1}{2} \langle b, (\gamma I - A)^{-1} b \rangle.
\end{equation}

\item \textbf{Metric coupling.}
From \eqref{eq:c_transform}, it is easy to check that 
$\Reg = -\Lambda \Leftrightarrow \Reg^\Phi = -\Lambda_C$
with $C = -\Phi$. From \cite[Proposition 6.1]{peyre_2019},
we have $\Lambda_C = -\Lambda$. Therefore, $\Reg^\Phi = -\Reg$.

\end{enumerate}

\subsection{Proofs for Proposition \ref{prop:properties_phi_fy_losses} 
(Properties of generalized Fenchel-Young losses)}
\label{proof:properties_phi_fy_losses}

\begin{enumerate}[topsep=0pt,itemsep=2pt,parsep=2pt,leftmargin=10pt]

\item \textbf{Non-negativity.} This follows immediately from the generalized
Fenchel-Young inequality.

\item \textbf{Zero loss.} If $p^\Phi_\Reg(v) = y$, then using
$L_\Reg^\Phi(v, y) 
= F(v, y) - F(v, p^\Phi_\Reg(v))$,
where $F(v, p) = \Reg(p) - \Phi(v, p)$,
we obtain $L_\Reg^\Phi(v, y) = 0$.
Let's now prove the reverse direction.
Since the maximum in \eqref{eq:phi_convex_conjugate}
exists, we have $F(v, p^\Phi_\Reg(v)) \le F(v, y)$.
If $L_\Reg^\Phi(v, y) = 0$, we have
$F(v, p^\Phi_\Reg(v)) = F(v, y)$.
Since the maximum is unique by assumption, this proves that 
$y = p^\Phi_\Reg(v)$.

\item \textbf{Gradient.} 
This follows directly from the definition \eqref{eq:phi_fy_loss}.

\item \textbf{Difference of convex (DC).} If $\Phi(v, p)$ is convex in $p$,
$\Reg^\Phi(v)$ is convex (Proposition
\ref{prop:properties_phi_convex_conjugate}).
Therefore, $v \mapsto \Reg^\Phi(v) - \Phi(v, p)$ for all $p \in \cC$.

\item \textbf{Smaller output set, smaller loss.}
    Let $\Reg \colon \cC \to \RR$ and let $\Reg'$ be the restriction of $\Reg$
    to $\cC' \subseteq \cC$, i.e., $\Reg'(p) \coloneqq \Omega + I_{\cC'}$, where
    $I_{\cC'}$ is the indicator function of $\cC'$. 
    From \eqref{eq:phi_convex_conjugate},
    we have $(\Reg')^\Phi(v) \le \Reg^\Phi(v)$ for all $v \in \cV$.
    From \eqref{eq:phi_fy_loss}, we therefore have 
    $L_{\Reg'}^\Phi(v, p) \le L_{\Reg}^\Phi(v, p)$ for all $v \in \cV$ 
    and all $p \in \cC'$.

\item \textbf{Quadratic lower-bound.}
If a function $F$ is $\gamma$-strongly convex over $\cC$ 
w.r.t.\ a norm $\|\cdot\|$, then
\begin{equation}
\frac{\gamma}{2} \|p - p'\|^2 \le F(p) - F(p') 
- \langle \nabla F(p'), p - p' \rangle
\label{eq:strongly_convex_def}
\end{equation}
for all $p, p' \in \cC$. 
If $\cC$ is a closed convex set,
we also have that 
$p^\star = \argmin_{p \in \cC} F(p)$ satisfies 
the optimality condition
\begin{equation}
\langle \nabla F(p^\star), p - p^\star \rangle \ge 0
\label{eq:constrained_optimum}
\end{equation}
for all $p \in \cC$
\cite[Eq.  (2.2.13)]{nesterov_2003}.
Combining \eqref{eq:strongly_convex_def} with $p' = p^\star$
and \eqref{eq:constrained_optimum}, we obtain
\begin{equation}
\frac{\gamma}{2} \|p - p^\star\|^2 \le F(p) - F(p^\star).
\end{equation}
Applying the above with $F(p) = \Reg(p) - \Phi(v, p)$
and using
$p^\star = p_\Reg^\Phi(v)$,
we obtain
\begin{equation}
\frac{\gamma}{2} \|y - p^\Phi_\Reg(v)\|^2 \le L_\Reg^\Phi(v, y).
\end{equation}

\item \textbf{Upper-bounds.}
If $F$ is $\alpha$-Lipschitz over $\cC$ with respect to a norm $\|\cdot\|$,
then for all $p,p' \in \cC$
\begin{equation}
|F(p) - F(p')| \le \alpha \|p - p'\|.
\end{equation}
Applying the above with $F(p) = \Reg(p) - \Phi(v, p)$
and $p' = p_\Reg^\Phi(v)$, we obtain
\begin{equation}
L^\Phi_\Reg(v, p) \le \alpha \|p - p^\Phi_\Reg(v)\|.
\end{equation}

If $\Phi(v, p)$ is concave in $p$, then its linear approximation always lies
above:
\begin{equation}
\Phi(v, p') 
\le 
\Phi(v, p) + \langle \nabla_2 \Phi(v, p), p' - p \rangle,
\end{equation}
for all $p,p' \in \cC$.
We then have
\begin{align}
L^\Phi_\Reg(v, p) 
&= \Reg^\Phi(v) + \Reg(p) - \Phi(v, p) \\
&= \max_{p' \in \cC} ~
\Phi(v, p') - \Reg(p') + \Reg(p) - \Phi(v, p) \\
&\le \max_{p' \in \cC} ~
\langle \nabla_2 \Phi(v, p), p' \rangle - \Reg(p')
+ \Reg(p) - \langle \nabla_2 \Phi(v, p), p \rangle \\
&= \Reg^*(\nabla_2 \Phi(v, p)) + \Reg(p) 
- \langle \nabla_2 \Phi(v, p), p \rangle \\
&= L_\Reg(\nabla_2 \Phi(v, p), p).
\end{align}

\end{enumerate}

\subsection{Proof of Proposition \ref{prop:calibration} (calibration)}
\label{proof:calibration}

\paragraph{Background.}

The pointwise target risk of $\widehat y \in \cY$ according to 
$q \in \triangle^{|\cY|}$ is
\begin{equation}
\ell(\widehat y, q) 
\coloneqq \EE_{Y \sim q} ~ L(\widehat y, Y).
\end{equation}
We also define the corresponding excess of pointwise risk,
the difference between the pointwise risk and the pointwise Bayes risk:
\begin{equation}
\delta\ell(\widehat y, q) 
\coloneqq \ell(\widehat y, q) - \min_{y' \in \cY} \ell(y', q).
\end{equation}
We can then write the risk of $f \colon \cX \to \cY$ 
in terms of the pointwise risk
\begin{align}
\cL(f) 
&\coloneqq \EE_{(X,Y) \sim \rho} ~ L(f(X), Y) \\
&= \EE_{X \sim \rho_\cX} \EE_{Y \sim \rho(\cdot | X)} L(f(X), Y) \\
&= \EE_{X \sim \rho_\cX} \ell(f(X), \rho(\cdot | X)).
\end{align}
Let us define the Bayes predictor 
\begin{equation}
f^\star \coloneqq \argmin_{f \colon \cX \to \cY} \cL(f).
\end{equation}
The Bayes risk is then
\begin{align}
\cL(f^\star) 
&= \min_{f \colon \cX \to \cY}
\EE_{X \sim \rho_\cX} \EE_{Y \sim \rho(\cdot | X)} L(f(X), Y) \\
&= \EE_{X \sim \rho_\cX} \min_{y' \in \cY} 
\EE_{Y \sim \rho(\cdot | X)} L(y', Y) \\
&= \EE_{X \sim \rho_\cX} \min_{y' \in \cY} \ell(y', \rho(\cdot | X)).
\end{align}
Combining the above, we can write the excess of risk 
of $f \colon \cX \to \cY$ as
\begin{equation}
\cL(f) - \cL(f^\star) 
= \EE_{X \sim \rho_\cX} \delta\ell(f(X), \rho(\cdot | X)).
\end{equation}

Similarly, with the generalized Fenchel-Young loss \eqref{eq:phi_fy_loss},
the pointwise surrogate risk of $v \in \cV$ according to 
$q \in \triangle^{|\cY|}$ is
\begin{equation}
\ell^\Phi_\Reg(v, q) 
\coloneqq \EE_{Y \sim q} ~ L^\Phi_\Reg(v, Y)
\end{equation}
and the excess of pointwise surrogate risk is
\begin{equation}
\delta\ell^\Phi_\Reg(v, q) 
\coloneqq \ell^\Phi_\Reg(v, q) - \min_{v' \in \cV} \ell^\Phi_\Reg(v', q).
\end{equation}
Let us define the surrogate risk of $g \colon \cX \to \cV$ by
\begin{equation}
\cL^\Phi_\Reg(g) 
\coloneqq \EE_{(X,Y) \sim \rho} ~ L^\Phi_\Reg(g(X), Y)
\end{equation}
and the corresponding Bayes predictor by
\begin{equation}
g^\star \coloneqq \argmin_{g \colon \cX \to \cV} \cL^\Phi_\Reg(g).
\end{equation}
We can then write the excess of surrogate risk of $g \colon \cX \to \cV$ as
\begin{equation}
\cL^\Phi_\Reg(g) - \cL^\Phi_\Reg(g^\star) 
= \EE_{X \sim \rho_\cX} \delta\ell^\Phi_\Reg(g(X), \rho(\cdot | X)).
\end{equation}
A calibration function \cite{steinwart_2007}
$\xi \colon \RR_+ \to \RR_+$ is a function relating the
excess of pointwise target risk and pointwise surrogate risk. 
It should be non-negative, 
convex and non-decreasing on $\RR_+$,
and satisfy $\xi(0) = 0$. Formally, given a decoder
$d \colon \cV \to \cY$, $\xi$ should satisfy
\begin{equation}
\xi(\delta\ell(d(v), q)) \le \delta\ell^\Phi_\Reg(v, q)
\label{eq:calibration_function}
\end{equation}
for all $v \in \cV$ and $q \in \triangle^{|\cY|}$.
By Jensen's inequality, this implies that
the target and surrogate risks are calibrated
for all $g \colon \cX \to \cV$ \cite{osokin_2017,nowak_2019}
\begin{equation}
\xi(\cL(d \circ g) - \cL(f^\star))
\le \cL^\Phi_\Reg(g) - \cL^\Phi_\Reg(g^\star).
\end{equation}
From now on, we can therefore focus on proving \eqref{eq:calibration_function}.

\paragraph{Upper-bound on the pointwise target excess risk.}

We now make use of the affine decomposition \eqref{eq:loss_affine_decomp}.
Let $\sigma \coloneqq \sup_{y \in \cY} \|V^\top \varphi(y)\|_*$, where
$\|\cdot\|_*$ denotes the dual norm of $\|\cdot\|$.
Let us define
\begin{equation}
\tilde{y}_L(u) 
\coloneqq \argmin_{\widehat y \in \cY}
\langle \varphi(\widehat y), V u + b \rangle
\end{equation}
and $\mu_\varphi(q) \coloneqq \EE_{Y \sim q}[\varphi(Y)] \in
\conv(\varphi(\cY)) \subseteq \varphi(\RR^k)$. 
From \cite[Lemma 2]{projection_oracles},
\begin{equation}
\delta\ell(\tilde{y}_L(u), q) \le 2 \sigma \|\mu_\varphi(q) - u\|
\quad \forall u \in \varphi(\RR^k), q \in \triangle^{|\cY|}.
\end{equation}
Using $y_L(p) = \tilde{y}_L(u)$ with $u = \varphi(p)$, we thus get
\begin{equation}
\delta\ell(y_L(p), q) \le 2 \sigma \|\mu_\varphi(q) - \varphi(p)\|
\quad \forall p \in \RR^k, q \in \triangle^{|\cY|}.
\label{eq:L_bayes_risk_bound}
\end{equation}

\paragraph{Bound on the pointwise surrogate excess risk (bilinear case).}

To highlight the difference with our proof technique, we first prove the result
in the bilinear case, assuming $\Reg$ is $\gamma$-strongly convex. 
We follow the same proof technique as \cite{nowak_2019,projection_oracles}
but unlike these works we do not require any Legendre-type assumption on $\Reg$.
If $\Phi(v, p) = \langle v, \varphi(p) \rangle = \langle v, p \rangle$, we have
\begin{align}
\ell^\Phi_\Reg(v, q) 
&= \EE_{Y \sim q} ~ L_\Reg(v, Y) \\
&= \Reg^*(v) + \EE_{Y \sim q} \Reg(Y) - \langle v, \mu(q) \rangle \\
&= \Reg^*(v) + \Reg(\mu(q)) - \langle v, \mu(q) \rangle 
+\EE_{Y \sim q} \Reg(Y) - \Reg(\mu(q)) \\
&= L_\Reg(v, \mu(q)) + \EE_{Y \sim q} \Reg(Y) - \Reg(\mu(q)),
\end{align}
where,
when $\varphi(y) = y$,
we denote 
$\mu(q) \coloneqq \EE_{Y \sim q}[Y] \in \conv(\cY) \subseteq \RR^k$
for short.
The quantity $\EE_{Y \sim q} \Reg(Y) - \Reg(\mu(q))$
is called Bregman information \cite{bregman_clustering} or Jensen gap,
and is non-negative if $\Reg$ is convex.
This term cancels out in the excess of pointwise surrogate risk
\begin{align}
\delta\ell_\Reg(v, q) 
\coloneqq \ell_\Reg(v, q) - \min_{v' \in \cV} \ell_\Reg(v', q) 
= L_\Reg(v, \mu(q)) - \min_{v' \in \cV} L_\Reg(v', \mu(q)).
\end{align}
Since the Fenchel-Young loss achieves its minimum at $0$, we have
\begin{align}
\delta\ell_\Reg(v, q) = L_\Reg(v, \mu(q)).
\end{align}
Therefore, the excess of pointwise surrogate risk can be written in
Fenchel-Young loss form.
By the quadratic lower-bound \eqref{eq:quadratic_lower_bound}
and the upper-bound \eqref{eq:L_bayes_risk_bound}, we have
\begin{align}
\delta\ell_\Reg(v, q) 
= L_\Reg(v, \mu(q)) 
\ge \frac{\gamma}{2} \|\mu(q) - p_\Reg(v)\|^2
\ge \frac{\gamma}{8 \sigma^2} \delta \ell(y_L(p_\Reg(v)), q))^2.
\end{align}
Therefore the calibration function 
with the decoder $d = y_L \circ p_\Reg$ is
\begin{equation}
\xi(\varepsilon) = \frac{\gamma \varepsilon^2}{8 \sigma^2}.
\end{equation}

\paragraph{Bound on the pointwise surrogate excess risk (linear-concave case).}

We now prove the bound assuming that $L_\Reg^\Phi$ is smooth and
$\Phi(v, p) = \langle v, \varphi(p) \rangle$.
This includes the
previous proof as special case because when $\Phi(v,p) = \langle v, p \rangle$,
then $L^\Phi_\Reg(v, y) = L_\Reg(v, y)$ is $\frac{1}{\gamma}$-smooth in $v$ if
and only if $\Reg(p)$ is $\gamma$-strongly convex in $p$ (cf.\
\S\ref{sec:background}).

By Theorem 4.22 in \cite{orabona_2019}, if a function $f(v)$ is $M$-smooth in
$v$ w.r.t. the dual norm $\|\cdot\|_*$ and is bounded below, then
\begin{equation}
f(v) - \min_{v' \in \cV} f(v')
\ge \frac{1}{2M} \|\nabla f(v)\|^2.
\end{equation}
With 
$f(v) = \ell^\Phi_\Reg(v, q) = \EE_{Y \sim q} ~ L_\Reg(v, Y)$,
which is non-negative,
we obtain
\begin{align}
\delta \ell^\Phi_\Reg(v,q)
&\coloneqq \ell^\Phi_\Reg(v, q) - \min_{v' \in \cV} \ell^\Phi_\Reg(v', q) \\
&\ge \frac{1}{2M} \|\nabla_1 \ell^\Phi_\Reg(v, q)\|^2 \\
&= \frac{1}{2M} \|\EE_{Y \sim q} \nabla_1 L^\Phi_\Reg(v, Y)\|^2 \\
&= \frac{1}{2M} \|\nabla \Reg^\Phi(v) 
- \EE_{Y \sim q} \nabla_1 \Phi(v, Y)\|^2 \\
&= \frac{1}{2M} \|\nabla_1 \Phi(v, p^\Phi_\Reg(v)) 
- \EE_{Y \sim q} \nabla_1 \Phi(v, Y)\|^2 \\
&= \frac{1}{2M} \|\varphi(p^\Phi_\Reg(v)) 
- \EE_{Y \sim q} \varphi(Y)\|^2 \\
&= \frac{1}{2M} \|\varphi(p^\Phi_\Reg(v)) 
- \mu_\varphi(q)\|^2.
\end{align}
We therefore have
\begin{align}
\delta\ell_\Reg^\Phi(v, q) 
\ge \frac{1}{2M} \|\mu_\varphi(q) - \varphi(p^\Phi_\Reg(v))\|^2
\ge \frac{1}{8 \sigma^2 M} \delta \ell(y_L(p_\Reg^\Phi(v)), q))^2.
\end{align}
Therefore the calibration function 
with the decoder $d = y_L \circ p^\Phi_\Reg$ is
\begin{equation}
\xi(\varepsilon) = \frac{\varepsilon^2}{8 \sigma^2 M}.
\end{equation}

\subsection{Proofs for Proposition \ref{prop:properties_phi_biconjugates} (Properties of
generalized biconjugates)}
\label{proof:properties_phi_biconjugates}

The proofs are similar to that for $C$-transforms \cite[Proposition
1.34]{santambrogio_2015}. 

\begin{enumerate}[topsep=0pt,itemsep=2pt,parsep=2pt,leftmargin=10pt]

\item \textbf{Lower bound.} 
Let $\Reg \colon \cC \to \RR$. We have
\begin{align}
\Reg^{\Phi \Psi}(p)
&\coloneqq \max_{v \in \cV} \Psi(p, v) - \Reg^\Phi(v) \\
&= \max_{v \in \cV} \Phi(v, p) - \Reg^\Phi(v) \\
&= \max_{v \in \cV} \Phi(v, p) - \left[\max_{p' \in \cC} \Phi(v, p') -
\Reg(p')\right] \\
&\le \max_{v \in \cV} \Phi(v, p) - \Phi(v, p) + \Reg(p) \\
&= \Reg(p).
\end{align}
Therefore, $\Reg^{\Phi \Psi}(p) \le \Reg(p)$ for all $p \in \cC$.
Analogously, 
for a function $\Lambda \colon \cV \to \RR$, we have
$\Lambda^{\Psi \Phi}(v) \le \Lambda(v)$ for all $v \in \cV$.

\item \textbf{Equality.} Since $\Reg$ is $\Psi$-convex,
there exits $\Lambda \colon \cV \to \RR$ such that $\Reg = \Lambda^\Psi$.
We then have $\Reg^\Phi = \Lambda^{\Psi \Phi}$.
Using the lower bound property, we get
$\Reg^\Phi(v) = \Lambda^{\Psi\Phi}(v) \le \Lambda(v)$ for all $v \in \cV$.
By the order reversing property, we have
$\Reg^{\Phi\Psi}(p) \ge \Lambda^\Psi(p) = \Reg(p)$ for all $p \in \cC$.
However, we also have $\Reg^{\Phi\Psi}(p) \le \Reg(p)$ for all $p \in \cC$.
Therefore, $\Reg^{\Phi\Psi}(p) = \Reg(p)$ for all $p \in \cC$.

\item \textbf{Tightest lower-bound.}
Let $\Reg'$ be any lower bound of $\Reg$ that is $\Psi$-convex.
Therefore, there exists $\Lambda \colon \cV \to \RR$ such that 
$\Reg'(p) = \Lambda^\Psi(p) \le \Reg(p)$ for all $p \in \cC$.
By the order reversing property, we have
$\Lambda^{\Psi\Phi}(v) \ge \Reg^\Phi(v)$ for all $v \in \cV$.
By the lower bound property, we also have
$\Lambda^{\Psi \Phi}(v) \le \Lambda(v)$ for all $v \in \cV$
and therefore, $\Lambda(v) \ge \Reg^\Phi(v)$ for all $v \in \cV$.
Applying the order reversing property once more, we get
$\Reg'(p) = \Lambda^\Psi(p) \le \Reg^{\Phi\Psi}(p)$ for all $p \in \cC$.
Therefore $\Reg^{\Phi\Psi}$ is the tightest lower bound of $\Reg$.

\end{enumerate}

\subsection{Proofs for Proposition \ref{prop:properties_Bregman} (Properties of
generalized Bregman divergences)}
\label{proof:properties_Bregman}

\begin{enumerate}[topsep=0pt,itemsep=2pt,parsep=2pt,leftmargin=10pt]

\item \textbf{Link with generalized Fenchel-Young losses.}
From \eqref{eq:theta_max} and \eqref{eq:theta_argmax}, we have
\begin{equation}
\Lambda^\bPhi(p) = \Phi(v_\Lambda^\bPhi(p), p) - \Lambda(v_\Lambda^\bPhi(p))
\quad \forall p \in \cC
\end{equation}
for any $\Lambda \colon \cV \to \RR$.
With $\Lambda = \Reg^\Phi$, if $\Reg$ is $\Phi$-convex, 
using Proposition \ref{prop:properties_phi_biconjugates}, we have
\begin{equation}
\Reg^{\Phi\bPhi}(p) 
= \Reg(p) 
= \Phi(v_\Lambda^\bPhi(p), p) - \Lambda(v_\Lambda^{\bPhi}(p))
\quad \forall p \in \cC.
\end{equation}
Plugging $\Reg(p')$ in \eqref{eq:phi_bregman_div}
and using the shorthand $v \coloneqq v_{\Reg^\Phi}^\bPhi(p')$,
we get
\begin{equation}
D_\Reg^\Phi(p, p') 
= \Reg(y) - \Phi(v, p) + \Reg^\Phi(v)
= L_\Reg^\Phi(v, p).
\end{equation}

\item \textbf{Non-negativity.} This follows directly from
the non-negativity of $L_\Reg^\Phi(v, p)$.

\item \textbf{Identity of indiscernibles}.
If $p = p'$, we immediately obtain $D_\Reg^\Phi(p, p') = 0$ from
\eqref{eq:phi_bregman_div}.
Let's prove the reverse direction.
If $D_\Reg^\Phi(p, p') = 0$,
we have $F(v_{\Reg^\Phi}^\bPhi(p'), p)
= F(v_{\Reg^\Phi}^\bPhi(p'), p')$ where
$F(v, p) = \Reg(p) - \Phi(v, p)$.
Since by assumption $F$ is strictly convex in $p$,
we obtain $p = p'$.

\item \textbf{Convexity.}
This follows from 
$D_\Reg^\Phi(p, p') = \Reg(p) - \Phi(v, p) + \text{const}$.

\item \textbf{Recovering Bregman divergences.}
If $\Phi(v, p) = \langle v, p \rangle$, 
we have
\begin{equation}
v_{\Reg^\Phi}^\bPhi(p)
= \argmax_{v \in \cV} \Phi(v, p) - \Reg^\Phi(v) 
= \argmax_{v \in \cV} \langle v, p \rangle - \Reg^*(v)
= \nabla \Reg(p).
\end{equation}
Plugging back in \eqref{eq:phi_bregman_div}, we obtain the Bregman divergence
\eqref{eq:bregman_div}.

\end{enumerate}

\newpage
\section{Experimental details}
\label{appendix:experimental_details}

\subsection{Multilabel classification}

\paragraph{Datasets.}

We used public datasets available at
\url{https://www.csie.ntu.edu.tw/~cjlin/libsvmtools/datasets/}.
Dataset statistics are summarized in Table \ref{table:multilabel_datasets}.
For all datasets, we normalize samples to have zero mean unit variance.  

\begin{table}[ht]
    \caption{Multilabel dataset statistics.}
    \label{table:multilabel_datasets}
    \small
    \centering
    \begin{tabular}{r c c c c c c c}
        \toprule
        Dataset & Type & Train & Dev & Test & Features &
        Classes & Avg. labels \\
        \midrule
        Birds & Audio & 134 & 45 & 172 & 260 & 19 & 1.96 \\
        Cal500 & Music & 376 & 126 & 101 & 68 & 174 & 25.98 \\
        Emotions & Music & 293 & 98 & 202 & 72 & 6 & 1.82 \\
        Mediamill & Video & 22,353 & 7,451 & 12,373 & 120 & 101 & 4.54 \\
        Scene & Images & 908 & 303 & 1,196 & 294 & 6 & 1.06\\
        %SIAM TMC & Text & 16,139 & 5,380 & 7,077 & 30,438 & 22 & 2.22\\
        Yeast & Micro-array & 1,125 & 375 & 917 & 103 & 14 & 4.17\\
        \bottomrule
    \end{tabular}
\end{table}

\paragraph{Experimental details.}

In all experiments, we set the activation $\sigma$ to 
$\text{relu}(a) \coloneqq \max\{0, a\}$.

\looseness=-1
For the unary model, we use a neural network with
one hidden-layer, i.e.,
$g_\theta(x) = W_2 \sigma(W_1 x + b_1) + b_2$, 
where $\theta = (W_2, b_2, W_1, b_1)$,
$W_2 \in \RR^{k \times m}$, $b_2 \in \RR^k$,
$W_1 \in \RR^{m \times d}$, $b_1 \in \RR^m$,
and $m$ is the number of hidden units.
We use the heuristic $m = \min\{100, d / 3\}$, where $d$ is the dimensionality
of $x$.

For the pairwise model, in order to obtain a negative semi-definite matrix $U$, 
we parametrize $U = -AA^\top$ with 
$A = [W_1 x + b_1, \dots, W_m x + b_m]$, 
where $W_j \in \RR^{k \times d}$ and $b_j \in \RR^k$
In our experiments, we choose a rank-one model, i.e., $m=1$.
Note that we use distinct parameters for the unary and pairwise models but
sharing parameters would be possible.

For the SPEN model, 
following \cite[Eq.\ 4 and 5]{belanger_2016},
we set the energy to $\Phi(v, p) = \langle u, p \rangle - \Psi(w, p)$,
where $v = (u, w)$, $u = g_\theta(x)$ and $w$ are the weights of the ``prior
network'' $\Psi$ (independent of $x$). 
We parametrize 
$\Psi(w, p) = W_2 \sigma(W_1 p + b_1) + b_2$, 
where $w = (W_2, b_2, W_1, b_1)$,
$W_2 \in \RR^{1 \times m}$, $b_2 \in \RR^1$ and
$W_1 \in \RR^{m \times k}$, $b_1 \in \RR^m$.
To further impose convexity of $\Psi$ in $p$, $W_2$ needs to be non-negative.
To do so without using constrained optimization,
we use the change of variable $W_2 = \text{softplus}(W'_2)$,
where $\text{softplus}(a) \coloneqq \log(1 + \exp(a))$ is used as an
element-wise bijective mapping.

\paragraph{Additional results.}

The gradient of $\Reg^\Phi(v)$ can be computed using the envelope theorem
(Proposition \ref{prop:properties_phi_convex_conjugate}), which does not require
to differentiate through $p^\Phi_\Reg(v)$.  Alternatively, since $\Reg^\Phi(v) =
\Phi(v, p^\Phi_\Reg(v)) - \Reg(p^\Phi_\Reg(v))$, we can also compute the
gradient of $\Reg^\Phi(v)$ by using the implicit function theorem,
differentiating through $p^\Phi_\Reg(v)$. To do so, we use the approach detailed
in \cite{blondel_implicit_diff}. Results in Table \ref{tab:multilabel_grad} show
that the envelope theorem performs comparably to the implicit function theorem,
if not slightly better.
Differentiating through $p^\Phi_\Reg(v)$ using the implicit function theorem
requires to solve a $k \times k$ system and its implementation is more
complicated than the envelope theorem. Therefore, we suggest to use the envelope
theorem in practice.

\begin{table}[h!]
\caption{Comparison of envelope and implicit function theorems on the pairwise
model (test accuracy in \%).}
\centering
\begin{tabular}{r c c c c c c}
\toprule
 & yeast & scene & mediamill & birds & emotions & cal500 \\
\midrule
Envelope theorem & 80.19 & {\bf 91.58} & {\bf 96.95} & {\bf 91.55} & {\bf
    80.56} & {\bf 85.73} \\
Implicit function theorem & {\bf 80.33} & {\bf 91.58} & {\bf 96.95} & 91.54 &
80.53 & 85.57 \\
\bottomrule
\end{tabular}
\label{tab:multilabel_grad}
\end{table}

In addition, we also compared the proposed generalized Fenchel-Young loss with
the energy loss, the binary cross-entropy loss and the generalized perceptron
loss, which corresponds to setting $\Reg(p) = 0$. As explained in
\S\ref{sec:energy_networks}, the cross-entropy loss requires to differentiate
through $p^\Phi_\Reg(v)$; we do so by implicit differentiation.
Table \ref{tab:multilabel_losses} shows that the generalized Fenchel-Young loss
outperforms these losses. As expected, the energy loss performs very
poorly, as it can only push the model in one direction \cite{lecun_2006}.
Using regularization $\Reg$, as advocated in this paper, is empirically
confirmed to be beneficial for accuracy.

\begin{table}[h!]
\caption{Comparison of loss functions for the pairwise model (accuracy in \%).}
\centering
\begin{tabular}{r c c c c c c}
\toprule
 & yeast & scene & mediamill & birds & emotions & cal500 \\
\midrule
Generalized FY loss & {\bf 80.19} & {\bf 91.58} & {\bf 96.95} & 91.55 & {\bf 80.56} & 85.73 \\
Energy loss & 42.35 & 33.02 & 40.92 & 14.29 & 55.50 & 39.27 \\
Cross-entropy loss & 79.00 & 90.78 & 96.77 & {\bf 91.56} & 78.08 & {\bf 85.89} \\
Generalized perceptron loss & 68.36 & 89.33 & 93.24 & 88.92 & 66.34 & 80.11 \\
\bottomrule
\end{tabular}
\label{tab:multilabel_losses}
\end{table}

\vspace{-0.5cm}
\subsection{Imitation learning}
\label{appendix:imitation_learning}

\paragraph{Experimental details.}

We run a hyperparameter search over the learning rate of the ADAM
optimizer, the number of
hidden units in the layers, the weight of the L2 parameters
regularization term and the scale of the energy regularization term $\Reg$. We
run the hyperparameter search for $4$ demonstration trajectories and select the
best performing ones based on the final performance (averaged over 3 seeds).

\begin{table}[ht]
\caption{Hyperparameter search for imitation learning.}
\centering
\begin{tabular}{r c c c c}
\toprule
Model & Learning rate & Params regularization & Energy regularization & Hidden units \\
& \{1e-4, 5e-4, 1e-3\} & \{0., 1., 10.\} & \{0.1, 1., 10.\} & \{16, 32, 64, 128\} \\
\midrule
Unary & 5e-4 & 0.0 & 1. & 16 \\
Pairwise & 1e-4 & 0.0 & 10. & 32 \\
\bottomrule
\end{tabular}
\end{table}

\paragraph{Environments.}

We also provide performance of the expert agent as detailed by Orsini et al.
\cite{orsini2021matters} as well as the
description of the observation and action spaces for each environment.
\begin{table}[h!]
\caption{Dimension of observation space, dimension of action space, expert
performance, and random policy performance for each environment.}
\centering
\begin{tabular}{l l l l l}
\toprule
Task & Observations & Actions & Random policy score & Expert score \\
\midrule
HalfCheetah-v2 & 17 & 6 & -282 & 8770 \\
Hopper-v2 & 11 & 3 & 18 & 2798 \\
Walker-v2 & 17 & 6 & 1.6 & 4118 \\
Ant-v2 & 111 & 8 & 123 & 5637 \\
\bottomrule\\
\end{tabular}
\label{tab:expert_perf}
\end{table}

\end{document}